\documentclass{article} 
\PassOptionsToPackage{table}{xcolor}
\usepackage[final]{colm2025_conference}

\usepackage{graphicx}
\usepackage{microtype}
\usepackage{hyperref}
\usepackage{url}
\usepackage{booktabs}
\usepackage{lineno}
\usepackage{todonotes}

\usepackage{algorithm}
\usepackage{algpseudocode}

\usepackage{amsmath}
\usepackage{multirow}
\usepackage{easyeqn}
\usepackage{bm}
\usepackage{amsfonts}
\usepackage[table]{xcolor}
\usepackage{tcolorbox}
\usepackage{listings}
\usepackage{wrapfig,lipsum,booktabs}
\lstdefinestyle{plain}{
    basicstyle=\fontsize{9}{10}\ttfamily,
    keywordstyle=\color{blue},
    commentstyle=\color{gray},
    stringstyle=\color{green},
    showstringspaces=false,
    breaklines=true,
    breakatwhitespace=false,
    breakindent=0pt,
    escapeinside={(*@}{@*)}
}
\lstdefinestyle{mystyle}{
    backgroundcolor=\color{white},   
    basicstyle=\ttfamily\small,       
    breaklines=true,                  
    frame=none,                       
    keywordstyle=\color{blue},        
    commentstyle=\color{gray},        
    stringstyle=\color{brown},        
    morekeywords={def, return},      
    escapeinside={\%*}{*)}           
}

\definecolor{o1}{HTML}{f4bb6e}
\definecolor{o2}{HTML}{f7f3e5}
\definecolor{o3}{HTML}{dadce0}

\usepackage{enumitem}
\usepackage{amssymb}
\usepackage{array}

\definecolor{darkblue}{rgb}{0, 0, 0.5}
\hypersetup{colorlinks=true, citecolor=darkblue, linkcolor=darkblue, urlcolor=darkblue}

\title{\includegraphics[width=1.2em, height=1.2em]{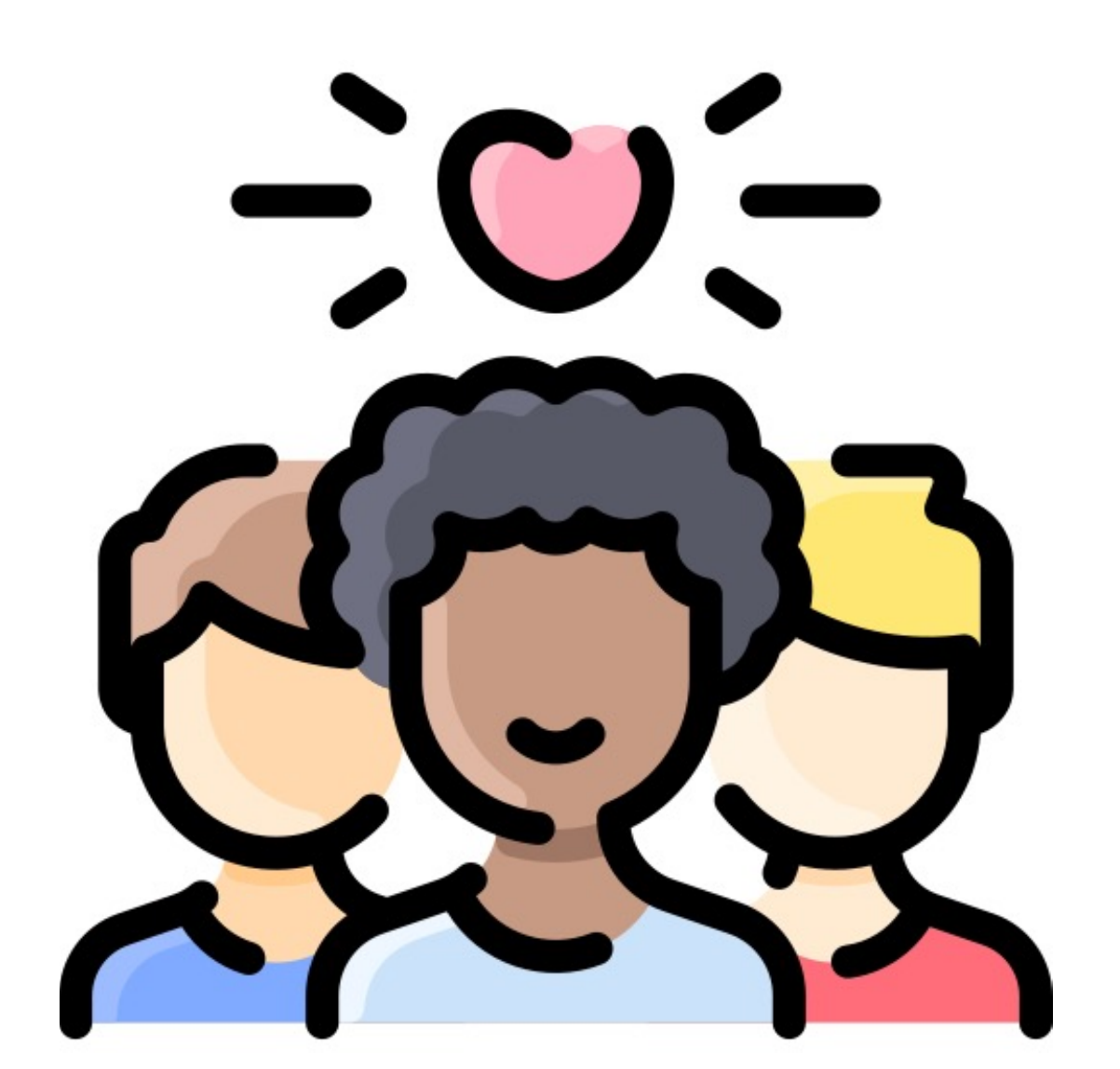} CultureCLIP: Empowering CLIP with Cultural Awareness through Synthetic Images and Contextualized Captions}

\author{Yuchen Huang$^{\mathbf{1}}$, ~Zhiyuan Fan$^{\mathbf{1}}$, ~Zhitao He$^{\mathbf{1}}$, ~Sandeep Polisetty$^{\mathbf{2}}$, 
~Wenyan Li$^{\mathbf{3}}$ \\ \textbf{Yi R. (May) Fung$^{\mathbf{1}}$}\\
$^{\mathbf{1}}$Hong Kong University of Science and Technology \\ $^{\mathbf{2}}$University of Massachusetts Amherst, 
$^{\mathbf{3}}$University of Copenhagen\\
\{yhuanggn, yrfung\}@cse.ust.hk \\
}

\NewDocumentCommand{\yi}
{ mO{} }{\textcolor{magenta}{\textsuperscript{\textit{may}}\textsf{\textbf{\small[#1]}}}}
\NewDocumentCommand{\zhiyuan}
{ mO{} }{\textcolor{red}{\textsuperscript{\textit{TODO:}}\textsf{\textbf{\small[#1]}}}}
\NewDocumentCommand{\zhitao}
{ mO{} }{\textcolor{cyan}{\textsuperscript{\textit{zhitao}}\textsf{\textbf{\small[#1]}}}}

\begin{document}

\ifcolmsubmission
\linenumbers
\fi

\maketitle

\begin{abstract}

Pretrained vision-language models (VLMs) such as CLIP excel in general multimodal comprehension but often struggle to capture nuanced, context-dependent visual cues. This makes it difficult to distinguish between similar-looking concepts with potentially different cultural meanings. Such deficiencies are mainly due to a limited amount of high-quality cultural data, contextual information, and the lack of negative examples that highlight subtle differences. To mitigate this, we design a data curation pipeline leveraging open-sourced VLMs and text-to-image models to construct \textbf{CulTwin}, a synthetic cultural dataset. This dataset consists of paired concept-caption-image triplets, where concepts visually resemble each other but are culturally different. Then, we fine-tune CLIP on CulTwin to develop \textbf{CultureCLIP}, which aligns cultural concepts with contextually enhanced captions and synthetic images through tailored contrastive learning. Experiments on culture-specific benchmarks show that CultureCLIP outperforms the base CLIP, achieving up to a notable 5.49\% improvement in fine-grained concept recognition on certain tasks while preserving CLIP's original generalization ability, validating the effectiveness of our data synthesis and VLM backbone training paradigm in capturing subtle cultural distinctions.\footnote{Our code is publicly available at \url{https://github.com/lukahhcm/CultureCLIP}.}
\end{abstract}
\section{Introduction}

\begin{figure*}[!t]
  \includegraphics[width=1\textwidth]{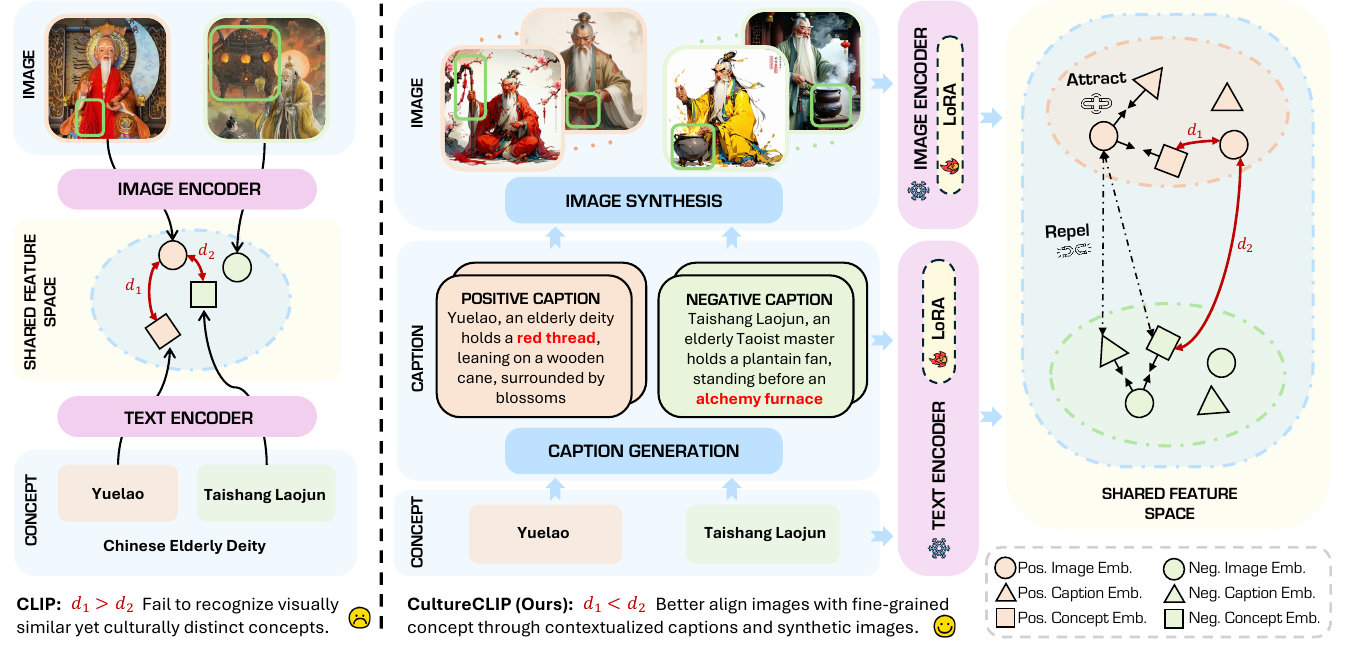}
  \vspace{-0.6em}
  \caption{CLIP vs. CultureCLIP. \textbf{Left: }The original CLIP model fails to capture fine-grained contextual visual cues (highlighted by the green box), leading to mismatches with cultural concepts. Although both the image and concept projections fall within the region outlined by the purple dashed box—i.e., the semantic space of Chinese elderly deities—the distance between the image of Yuelao (pink circle) and its correct concept (pink square) is greater than that to an incorrect one such as Taishang Laojun (blue square), i.e., $d_1 > d_2$. \textbf{Right: }CultureCLIP improves fine-grained cultural understanding by jointly aligning concepts with contextualized captions and their corresponding synthetic images, while repelling unrelated concepts and captions in the embedding space (see details in Section \ref{sec:cultureclip}). }
  \label{fig:teaser_cultureclip}
\end{figure*}

Recent advancements in vision-language reasoning \citep{zhang2024can, lu2024deepseek} have revolutionized multimodal understanding by efficiently integrating visual and linguistic semantics within a shared feature space~\citep{jia2021scalingvisualvisionlanguagerepresentation,radford2021learning}. By leveraging large-scale image-text corpora, models such as Contrastive Language-Image Pretraining (CLIP) \citep{radford2021learning} exhibit remarkable generalization capabilities across diverse downstream tasks, including visual question answering (VQA)\citep{shen2021clipbenefitvisionandlanguagetasks,song-etal-2022-clip}, cross-modal retrieval \citep{koukounas2024jinaclipclipmodel,baldrati2023composedimageretrievalusing}, and zero-shot image classification \citep{Zhou_2022,saha2024improvedzeroshotclassificationadapting}. CLIP is built on a contrastive learning objective, where two separate encoders are trained to bring matching image-text pairs closer in the feature space while pushing apart non-matching pairs within the same batch. Due to the concise nature of the text in its training data, CLIP is effective at coarse-grained semantic alignment, particularly in identifying the general type of the main object \citep{radford2021learning, zhang2024long}. However, it often struggles with fine-grained alignment, especially when contextual visual details, such as specific accessories, stylistic cues, or symbolic elements that convey meaning only within particular cultural contexts, are required to distinguish between visually similar but culturally distinct concepts. 

For example, CLIP might correctly identify both \textit{Yuelao} (the Chinese god of love and marriage) and \textit{Taishang Laojun} (a Daoist patriarch) as elderly Chinese deities, but it often struggles to tell them apart because of its insensitivity to capturing subtle yet crucial visual details, like the \textit{red thread} for \textit{Yuelao}, which stands for love, or the \textit{alchemy furnace} for \textit{Taishang Laojun}, which stands for immortality (both shown in green in Figure \ref{fig:teaser_cultureclip}). These culturally specific visual cues, however, are exactly what people in particular cultural groups use to distinguish fine-grained concepts. This raises an intriguing question: \textbf{How can we teach CLIP to capture such details so that it can differentiate between cultural concepts that share visual similarities?} 

A natural approach is to curate a large-scale dataset comprising visually similar cultural concept pairs (i.e., original concepts alongside their hard negatives) accompanied by image-text pairs enriched with contextual cultural knowledge and illustrating subtle visual distinctions. However, collecting such data presents three major challenges: \textbf{First, high-quality cultural image-text pairs are scarce and costly to annotate.} Existing manually curated cultural datasets are typically limited in scale, often containing only a few thousand samples \citep{nikandrou2024crope, bhatia2024local, romero2024cvqaculturallydiversemultilingualvisual}, due to the labor-intensive nature of data collection and annotation, as well as the need for domain-specific expertise. While web-scraped data might serve as an alternative \citep{xu2023demystifying}, it frequently introduces substantial noise: images may be ethically inappropriate, copyright-protected, low-resolution, or culturally misleading, and the corresponding text may be mismatched or entirely absent. \textbf{Second, CLIP's original training strategy inherently favors concise captions.} Specifically, it imposes a 77-token limit, with the majority of alignment achieved within the first 20 tokens \citep{zhang2024long}. This design makes it particularly challenging to incorporate lengthy, information-rich cultural context directly into the text for training, as doing so may disrupt the original image-text alignment learned by the model. \textbf{Third, fine-grained hard negatives specifically tailored for visually similar concepts are lacking.} In the original CLIP framework, negative samples are randomly drawn within each batch, which significantly reduces the likelihood of including conceptually similar but visually distinct examples. Although recent works \citep{yuksekgonul2022and, patel2024tripletclip} introduce harder negatives by making minor modifications to captions or images, they still lack structured negative samples that highlight both coarse-grained similarities and fine-grained differences, which are crucial for enabling culturally grounded visual distinctions.

Motivated by these challenges, we introduce \textbf{CulTwin}, a synthetic dataset of paired concept-caption-image triplets, which we call \textit{Twin Cards}.  In each card, two similar concepts are paired together, and their captions are enriched with cultural background knowledge using a vision-language model. The corresponding images are then generated from these captions by a text-to-image model. Building on CulTwin, we propose \textbf{CultureCLIP}, a contrastive learning framework that jointly aligns concepts, captions, and images in a shared embedding space. As illustrated in Figure~\ref{fig:teaser_cultureclip} (right), our training objective works like a magnetic field: it attracts each concept toward its corresponding caption and image, while repelling it from the captions and images of its culturally contrasting counterpart. Experiments on culture-specific benchmarks show that CultureCLIP significantly outperforms the original CLIP model, achieving over 5\% improvement in fine-grained concept recognition on specific tasks. These results highlight the effectiveness of our synthetic dataset and training methodology in capturing subtle cultural nuances.
\section{Related Work}

\paragraph{Advancements in Vision-Language Models} Recent studies on multimodal reasoning \citep{peng2025lmmr1empowering3blmms, zhao2025r1omniexplainableomnimultimodalemotion, zhang2025vlm2benchcloserlookvlms,he2025mmboundaryadvancingmllmknowledge} have significantly advanced vision-language models and broadened their capabilities across various downstream tasks and domains. For instance, Med-Flamingo \citep{moor2023med} unlocks medical VQA abilities through continued pre-training on paired and interleaved medical image-text data, while ChemVLM \citep{li2024chemvlm}, trained on bilingual image-text data, enhances the joint understanding of textual and visual chemical information. In the cultural domain, CultureVLM \citep{liu2025culturevlm} improves cultural understanding by fine-tuning on a large-scale multimodal benchmark, CultureVerse. Despite these advancements, existing vision-language models still struggle to capture fine-grained visual cues and often misclassify visually similar but culturally distinct concepts. We propose CultureCLIP, which aligns cultural concepts with enriched captions and synthetic images through contrastive learning, improving cultural differentiation while preserving generalization capabilities.

\paragraph{Data for Vision-Language Pre-training} Cross-modal mutual information maximization relies on large-scale, diverse training data that captures real-world concepts and relationships. With more than 5 billion internet-derived image-caption pairs, the LAION dataset \citep{schuhmann2022laion} serves as a critical training resource. MetaCLIP \citep{xu2023demystifying} formalizes CLIP's implicit data selection via explicit metadata balancing, creating 400M CommonCrawl pairs. SynthCLIP \citep{hammoud2024synthclip} reduces reliance on web-scraped data by generating over 30 million synthetic pairs. LaCLIP \citep{fan2023improving} enhances text augmentation through in-context language model rewriting. For human-centric AI, fine-grained cultural understanding is essential, yet culturally relevant multimodal data is scarce. While \cite{nayak2024benchmarking, liu2025culturevlm} introduced cultural datasets, the diversity of images hinders vision-language models from learning cultural distinctions. In this work, we construct CulTwin, a synthetic dataset comprising concept-caption-image triplets enriched with cultural contextual knowledge.


\paragraph{Contrastive Pre-training} Contrastive learning has become a strong method for multimodal representation learning, with CLIP \citep{radford2021learning} demonstrating scalability and zero-shot transfer potential. More efficient contrastive pre-training methods have been proposed for finer-grained multimodal representations learning
\citep{zhang2024contrasting, patel2024tripletclip}. 
BLIP-2 \citep{li2023blip} introduces a lightweight Querying Transformer for cost-efficient pre-training.  
NegCLIP \citep{yuksekgonul2022and} generates hard negative captions through semantic perturbations. TripletCLIP \citep{patel2024tripletclip} uses hard negative pairs with a triplet contrastive loss. Existing contrastive pre-training methods focus on caption-image pairs and their negative samples, but fail to capture culturally relevant information due to its multidimensional nature. To address this, CultureCLIP enhances cultural understanding by aligning cultural concepts with contextualized captions and synthetic images, while separating unrelated concepts in the embedding space.

\section{CulTwin: A Three-stage Cultural Data Curation Pipeline}
\begin{figure*}[!t]
  \includegraphics[width=1\textwidth]{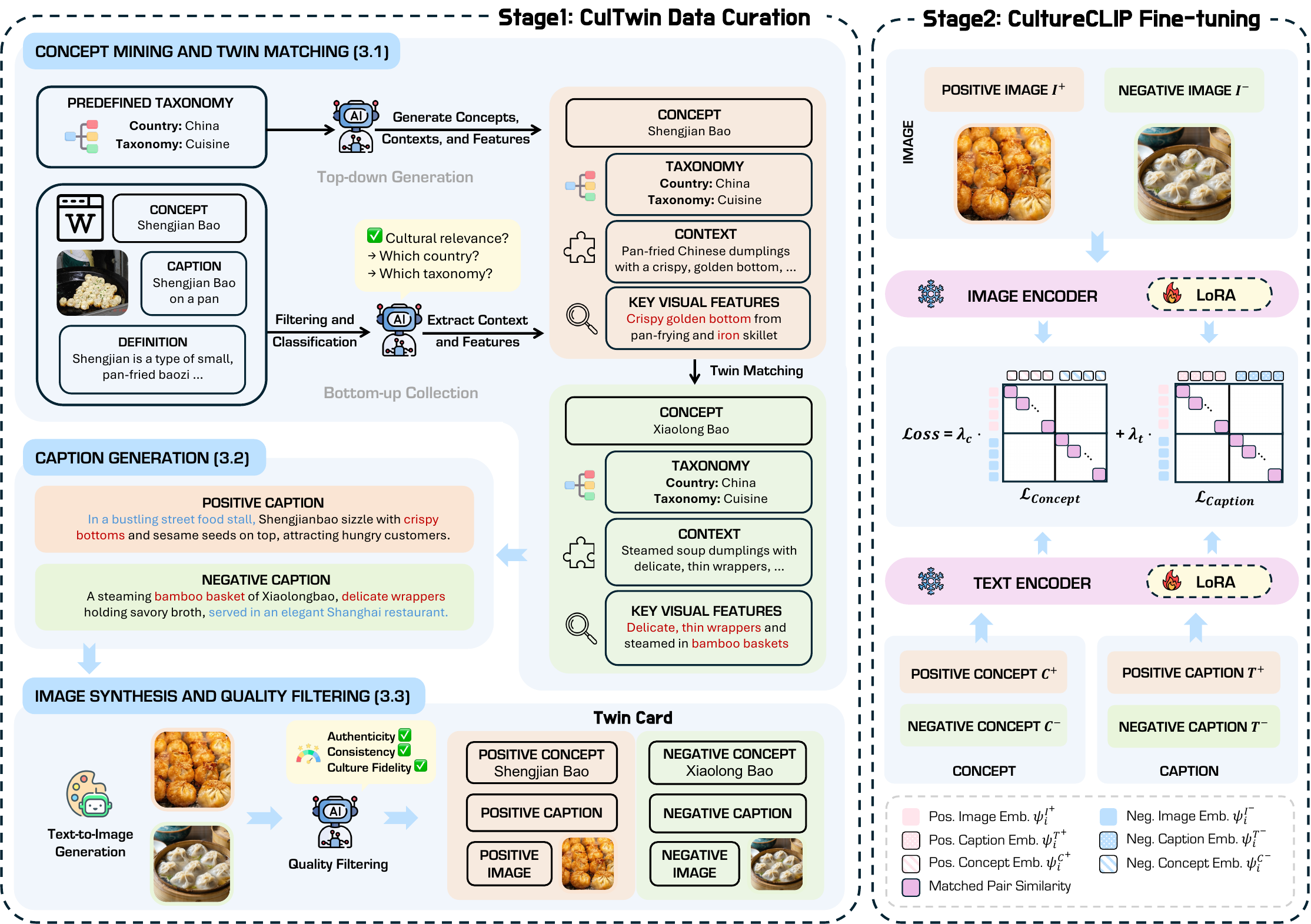}
  \vspace{-0.6em}
  \caption{\textbf{Left:} Data curation pipeline in CulTwin. \textbf{Right:} Architecture of CultureCLIP.}
  \label{fig:cultwin_pipeline}
\end{figure*}

High-quality and diverse data has been shown to be essential for training models like CLIP \citep{nguyen2023qualityquantityinteractiondataset, fang2022datadeterminesdistributionalrobustness}. In this section, we present a three-stage data curation pipeline for constructing \textbf{CulTwin}, a synthetic cultural dataset composed of \textit{Twin Cards}—pairs of concept-caption-image triplets that are visually similar but culturally distinct. The pipeline begins by collecting culturally grounded concepts, their background knowledge, and visual features, and performing twin matching to identify negative samples that are visually similar but culturally different (Section~\ref{sec:stage1}). Next, diverse captions are generated by leveraging cultural context and key visual features through a large language model (LLM) (Section~\ref{sec:stage2}). Finally, images are synthesized from the captions and evaluated using a Vision-Language Model (VLM), which scores each image based on authenticity, consistency, and cultural fidelity to guide data quality filtering (Section~\ref{sec:stage3}). The resulting concept-caption-image triplets are then organized into \textit{Twin Cards} for later fine-grained contrastive training. Figure~\ref{fig:cultwin_pipeline} (Left) provides an overview of the full CulTwin data curation pipeline.

\subsection{Concept Mining and Twin Matching}
\label{sec:stage1}
We begin with a manually predefined taxonomy covering 229 countries and 8 cultural categories, including \textit{Cuisine, Clothing, Animals \& Plants, Art, Architecture, Daily Life, Symbols, and Festivals}. This taxonomy is designed to capture a broad and representative set of cultural elements (definitions of each category are provided in Appendix~\ref{app:taxonomy}). We then collect culturally meaningful concept candidates through both bottom-up collection and top-down generation.

\paragraph{Bottom-up Concept Collection} In this approach, we first collect candidate concepts and their background information (definitions, images, captions) from Wikipedia. We then use Qwen2.5-VL \citep{bai2025qwen25vltechnicalreport} to assess each concept's cultural relevance, discarding any that are not strongly related to the predefined categories. To ensure strict filtering, we designed prompts (Appendix~\ref{app:prompts}) that favor rejection over uncertain inclusion, prioritizing data quality over quantity. For the retained concepts, we assign metadata (country and cultural category) and extract key contextual and visual features.

\paragraph{Top-down Concept Generation} In this approach, we further expand the concept pool using the predefined taxonomy. For each cultural category and country, Qwen2.5-VL is employed to generate culturally grounded concepts, along with associated context and visual features.

After obtaining these filtered concepts from both approaches, we perform \textit{twin matching} to identify culturally distinct but visually similar hard negatives for each concept, using Qwen2.5-VL conditioned on its context and key visual features. These final concept pairs, together with their metadata and contextual information, form the foundation for generating detailed captions and images in subsequent stages.

\subsection{Diverse Caption Generation}
\label{sec:stage2}
In the second stage, we generate diverse, culturally rich captions for each concept in the paired sets. These captions are designed to highlight both key visual elements and cultural nuances, preserving subtle distinctions before image synthesis. We leverage Qwen2.5-VL to incorporate cultural context and salient visual features, producing contextualized descriptions. To mitigate the risk of limited diversity and potential overfitting in synthetic data \citep{hammoud2024synthclip}, we guide the model to vary aspects such as artistic style, scene setting, and compositional details, thereby enriching the visual representation of each concept.

\subsection{Image Synthesis and Quality Filtering} 
\label{sec:stage3}

\begin{table}[t]
\centering
\caption{Filtering Outcomes and Image Quality Scores Across Diverse Cultural Taxonomies}
\label{tab:quality_evaluation}
\resizebox{\linewidth}{!}{
\begin{tabular}{lcccccccc}
\toprule
\multirow{3}{*}{\textbf{Taxonomy}} & 
\multicolumn{6}{c}{\textbf{Image Quality Scores After Filtering}} & 
\multicolumn{2}{c}{\textbf{Filtering Outcomes}} \\
\cmidrule(lr){2-7} \cmidrule(lr){8-9}
& 
\multicolumn{3}{c}{\textbf{MLLM-as-a-Judge}} & 
\multicolumn{3}{c}{\textbf{Human Evaluation}} & 
\multirow{2}{*}{\textbf{Pass \%}} & 
\multirow{2}{*}{\textbf{Retained}} \\
\cmidrule(lr){2-4} \cmidrule(lr){5-7}
& 
\multicolumn{1}{c}{\textbf{Auth $\uparrow$}} & 
\multicolumn{1}{c}{\textbf{Cons $\uparrow$}} & 
\multicolumn{1}{c}{\textbf{Fid $\uparrow$}} &

\multicolumn{1}{c}{\textbf{Auth $\uparrow$}} & 
\multicolumn{1}{c}{\textbf{Cons $\uparrow$}} & 
\multicolumn{1}{c}{\textbf{Fid $\uparrow$}} &
& \\
\midrule
\textbf{Cuisine} & 
\cellcolor{o2}4.403 ± 0.528 & \cellcolor{o3}3.548 ± 0.627 & \cellcolor{o3}3.750 ± 0.719 & \cellcolor{o2}4.517 ± 0.866 & 3.533 ± 0.957 & 3.550 ± 1.023 & \cellcolor{o2}77.46 & 24824/32,046 \\
\textbf{Clothing} & 
\cellcolor{o1}4.469 ± 0.397 & 3.862 ± 0.422 & 4.002 ± 0.152 & \cellcolor{o1}4.550 ± 0.956 & 3.483 ± 1.218 & 3.433 ± 1.202 & 75.52 & 7250/9,600 \\
\textbf{Animal \& Plants} & 
4.350 ± 0.415 & \cellcolor{o2}3.979 ± 0.528 & \cellcolor{o1}4.284 ± 0.366 & 4.183 ±1.162 & 3.783 ± 1.185 & 3.800 ± 1.137 & 75.63 & 10966/14,500 \\
\textbf{Art} & 
4.390 ± 0.444 & \cellcolor{o1}4.032 ± 0.415 & 4.093 ± 0.253 & 4.433 ± 0.761 & \cellcolor{o3}2.950 ± 1.189 & \cellcolor{o3}2.917 ± 1.144 & 72.67 & 15806/21,750 \\
\textbf{Architecture} & 
4.222 ± 0.349 & 3.897 ± 0.402 & 3.997 ± 0.335 & 4.183 ± 1.025 & 3.117 ± 1.112 & \cellcolor{o3}2.917 ± 1.100 & \cellcolor{o3}61.32 & 6377/10,400 \\
\textbf{Daily Life} & 
4.239 ± 0.436 & 3.882 ± 0.489 & \cellcolor{o2}4.152 ± 0.308 & 4.483 ± 0.904 & \cellcolor{o1}4.050 ± 1.132 & \cellcolor{o1}4.100 ± 1.179 & 72.80 & 6188/8,500 \\
\textbf{Symbol} & 
\cellcolor{o3}3.802 ± 0.371 & 3.830 ± 0.291 & 3.977 ± 0.109 & 4.467 ± 0.718 & \cellcolor{o2}3.950 ± 1.310 & \cellcolor{o2}3.833 ± 1.227 & \cellcolor{o1}85.12 & 681/800 \\
\textbf{Festival} & 
3.971 ± 0.326 & 3.921 ± 0.332 & 3.994 ± 0.061 & \cellcolor{o3}4.167 ± 0.986 & 3.533 ± 1.087 & 3.533 ± 1.008 & 72.12 & 1731/2,400 \\
\bottomrule
\end{tabular}
}
\vspace{0.2cm}
\parbox{\textwidth}{
\footnotesize
Scores are averaged over three dimensions—\textbf{Auth} (image authenticity), \textbf{Cons} (concept consistency), and \textbf{Fid} (cultural fidelity)—rated from 1 to 5. \textbf{Pass \%} indicates the proportion of images passing quality thresholds; \textbf{Retained} shows the number of remaining samples per category.
}
\end{table}

\begin{figure*}[!t]
  \includegraphics[width=1\textwidth]{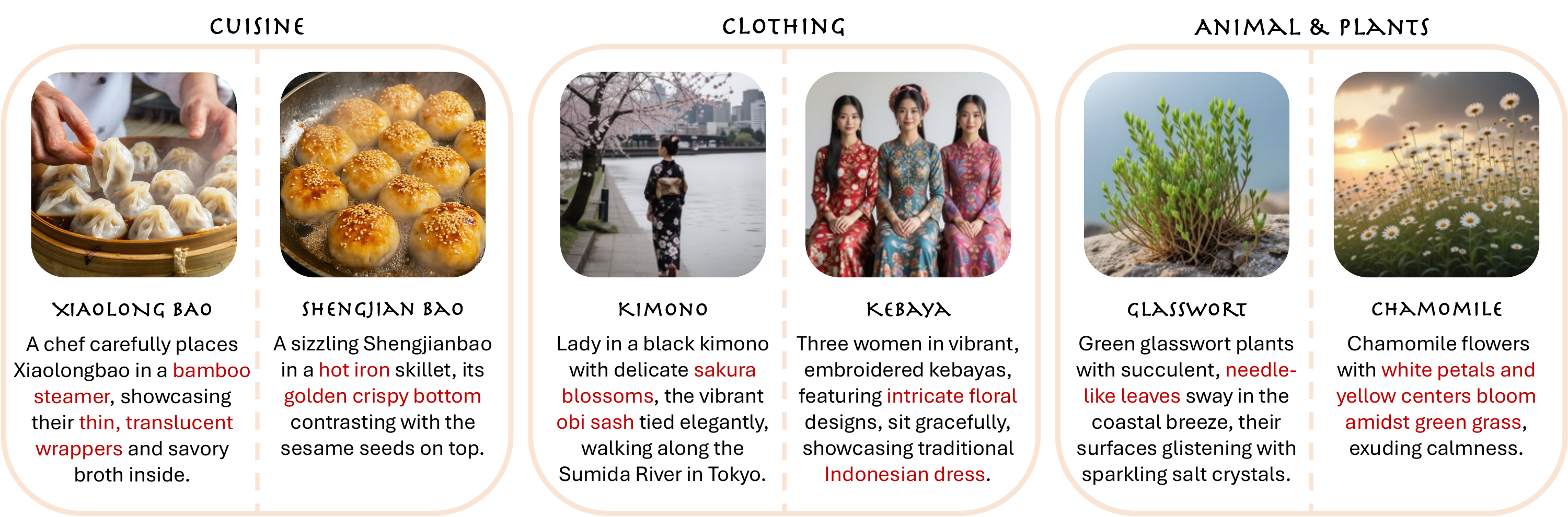}
  \vspace{-1.1em}
    \caption{\textit{Twin Card} examples from the \textit{Cuisine}, \textit{Clothing}, and \textit{Animals \& Plants} categories. Each Twin Card contrasts two culturally distinct yet visually similar concepts, with each side comprising a concept-caption-image triplet (see Section~\ref{sec:stage3}). Additional examples are provided in Appendix~\ref{app:cultwin details}.}
  \label{fig:examples}
\end{figure*}

After caption generation, we synthesize images using Stable Diffusion 3.5 \citep{rombach2022highresolutionimagesynthesislatent}, followed by quality filtering with MLLM-as-a-Judge \citep{chen2024mllmasajudgeassessingmultimodalllmasajudge}, implemented using Qwen-VL-2.5. Each synthesized image is evaluated across three key dimensions on a scale from 1 to 5:

\textbf{1) Authenticity:} Evaluates the physical realism and adherence to common human understanding. 
\textbf{2) Consistency:} Assesses alignment between the image and its caption, ensuring accurate concept representation. 
\textbf{3) Cultural Fidelity:} Examines the preservation and correctness of cultural features specific to the concept.

Images receiving a score of 1 in any dimension or an average score below 3 are discarded. To validate the automated filtering, we also perform a human evaluation on a sampled subset, where three PhD-level experts independently score the images using the same criteria. Summary statistics of the automated and human evaluations are shown in Table~\ref{tab:quality_evaluation}. 

Categories like \textit{Cuisine} and \textit{Clothing} exhibit high \textbf{Authenticity} scores, reflecting their tangible and universally recognized nature, making them easier to evaluate accurately by both models and human judges. In contrast, more abstract categories such as \textit{Festival} tend to have lower \textbf{Authenticity} and \textbf{Cultural Fidelity} scores, particularly in human evaluations, due to their complex and diverse cultural elements. Furthermore, \textit{Art} and \textit{Architecture} show a noticeable gap between automated and human evaluations, especially in \textbf{Consistency} and \textbf{Cultural Fidelity}. These categories involve nuanced cultural and conceptual details that automated models struggle to capture, highlighting the importance of human judgment for these intricate domains.

Each final \textit{Twin Card} is constructed by assembling two triplets—each consisting of a concept, its caption (from Section~\ref{sec:stage2}), and the corresponding synthesized image—to explicitly highlight cultural contrast while maintaining visual similarity. Example Twin Cards are illustrated in Figure~\ref{fig:examples}, and additional details of CulTwin can be found in Appendix~\ref{app:cultwin details}.
\section{CultureCLIP: Fine-Grained Cultural Alignment}  
\subsection{Preliminary}

CLIP learns joint image-text representations via an image encoder \(\mathcal{F}: \mathcal{I} \rightarrow \mathbb{R}^d\) and a text encoder \(\mathcal{G}: \mathcal{T} \rightarrow \mathbb{R}^d\), projecting inputs into a shared embedding space \(\mathcal{V}\) of dimension \(d\). Given a batch of \(N\) image-text pairs \(\{(I_i, T_i)\}_{i=1}^N\), representations are computed as \(\psi_i^I = \mathcal{F}(I_i)\) and \(\psi_i^T = \mathcal{G}(T_i)\). CLIP constructs a similarity matrix \(S \in \mathbb{R}^{N \times N}\), where each entry \(S_{i,j} = \text{sim}(\psi_i^I, \psi_j^T)\) denotes the cosine similarity between image \(I_i\) and text \(T_j\). The contrastive loss encourages each aligned pair (on the diagonal of \(S\)) to have a higher similarity than all mismatched pairs in the same row or column. Formally:
\small
\begin{equation}
\mathcal{L}_{\text{CLIP}} (I, T)= \mathcal{L}_{\text{I2T}}(I, T) + \mathcal{L}_{\text{T2I}}(I, T),
\end{equation}
\small
\begin{equation}
\mathcal{L}_{\text{I2T}}(I, T) = -\frac{1}{N} \sum_{i=1}^{N} \log 
\frac{\exp(\text{sim}(\psi_i^I, \psi_i^T)/\tau)}
{\sum_{k=1}^{N} \exp(\text{sim}(\psi_i^I, \psi_k^T)/\tau)},
\end{equation}

\small
\begin{equation}
\mathcal{L}_{\text{T2I}}(I, T) = -\frac{1}{N} \sum_{i=1}^{N} \log 
\frac{\exp(\text{sim}(\psi_i^I, \psi_i^T)/\tau)}
{\sum_{k=1}^{N} \exp(\text{sim}(\psi_k^I, \psi_i^T)/\tau)}.
\end{equation}
Here, \(\tau\) is a temperature parameter that controls the sharpness of the similarity distribution.

NegCLIP~\citep{yuksekgonul2022and} builds on this framework by introducing hard negative captions \(T^-\), derived through semantic perturbations of the original texts \(T^+\). These hard negatives are added alongside the standard in-batch negatives, forming an extended candidate set \(\tilde{T} = \{T^+\} \cup \{T^-\}\) and resulting in a similarity matrix \(\tilde{S} \in \mathbb{R}^{N \times 2N}\). The contrastive objective is thus extended beyond standard in-batch negatives to also include explicitly constructed hard negatives:
\small
\begin{equation}
\mathcal{L}_{\text{NegCLIP}} (I, T^+, T^-)= \mathcal{L}_{\text{I2T\_neg}}(I, T^+, T^-) + \mathcal{L}_{\text{T2I}}(I, T^+),
\end{equation}
\small
\begin{equation}
\mathcal{L}_{\text{I2T\_neg}}(I, T^+, T^-) = -\frac{1}{N} \sum_{i=1}^{N} \log 
\frac{\exp(\text{sim}(\psi_i^I, \psi_i^{T^+})/\tau)}
{\sum_{k=1}^{N} \exp(\text{sim}(\psi_i^I, \psi_k^{T^+})/\tau) + \sum_{m=1}^{N} \exp(\text{sim}(\psi_i^I, \psi_m^{T^-})/\tau)}.
\end{equation}

TripletCLIP~\citep{patel2024tripletclip} further incorporates hard negative images \(I^-\). The training objective encourages \(I^+\) to be closer to \(T^+\) than to \(T^-\), and symmetrically, \(I^-\) to align more with \(T^-\) than with \(T^+\). This is achieved by summing two NegCLIP-style losses:
\small
\begin{equation}
\mathcal{L}_{\text{TripletCLIP}}(I^+, I^-, T^+, T^-) = 
\mathcal{L}_{\text{NegCLIP}}(I^+, T^+, T^-) + 
\mathcal{L}_{\text{NegCLIP}}(I^-, T^-, T^+).
\end{equation}

While prior work enhances image-text alignment using modality-specific hard negatives, it tends to focus on coarse-grained semantic differences. We extend this framework by introducing abstract concepts as anchors to better connect detailed captions and images. Through a refined training objective (Section~\ref{sec:cultureclip}), our model intends to capture fine-grained cultural semantics with greater precision.

\subsection{CultureCLIP}
\label{sec:cultureclip}
To better capture fine-grained cultural semantics, we build on the \textit{Twin Cards} introduced in \textbf{Cultwin}, where each card contains two triplets—\((C^+, T^+, I^+)\) and \((C^-, T^-, I^-)\)—representing visually similar but culturally distinct concepts. These pairs serve as hard negatives for each other. Our goal is to align each concept with its corresponding caption and image while distinguishing it from its cultural counterpart.

To achieve this, we propose \textbf{CultureCLIP}, a contrastive learning framework that jointly embeds concepts, captions, and images into a shared semantic space. As illustrated in Figure~\ref{fig:teaser_cultureclip}, we encourage semantic \textbf{attraction} within each triplet and \textbf{repulsion} across its cultural counterpart. In addition, the overall architecture of our pipeline, including this learning scheme, is shown in Figure~\ref{fig:cultwin_pipeline} (Right). We employ a shared text encoder for both concepts and captions to preserve the original alignment between images and captions while aligning concepts with images. This shared encoder design ensures that fine-tuning for cultural distinctions does not degrade the model’s general cross-modal alignment ability. Our overall training objective is defined as:
\small
\begin{equation}
\mathcal{L}_{\text{CultureCLIP}} = \lambda_c \cdot \mathcal{L}_{\text{concept}} + \lambda_t \cdot \mathcal{L}_{\text{caption}},
\end{equation}
where \(\lambda_c\) and \(\lambda_t\) balance the contributions of concept-level and caption-level objectives. Both objectives are symmetrically formulated to promote attraction within positive triplets and repulsion from their cultural counterparts:
\small
\begin{equation}
\mathcal{L}_{\text{caption}} = \mathcal{L}_{\text{NegCLIP}}(I^+, T^+, T^-) + \mathcal{L}_{\text{NegCLIP}}(I^-, T^-, T^+),
\end{equation}
\small
\begin{equation}
\mathcal{L}_{\text{concept}} = \mathcal{L}_{\text{NegCLIP}}(I^+, C^+, C^-) + \mathcal{L}_{\text{NegCLIP}}(I^-, C^-, C^+),
\end{equation}
This structure jointly anchors abstract cultural concepts to specific visual-textual cues, enhancing the model's ability to differentiate subtle cultural semantics. To further preserve the original model's generalization ability, we apply the parameter-efficient LoRA method for fine-tuning on this loss, rather than directly training both the visual and text encoders.

\section{Experiment}
\subsection{Experimental Setup}

\paragraph{Model Choices and Prompting Strategies} In this work, we leverage Qwen2.5-VL \citep{bai2025qwen25vltechnicalreport} as our LLM/VLM (note that the specific choice of model is not the primary focus of this paper). Although more powerful LLMs could potentially further improve data quality, we consider Qwen2.5-VL to strike a good balance between performance and cost, making it a practical choice for our experiments. Given the significant impact that prompt design has on LLM performance \citep{brown2020languagemodelsfewshotlearners}, we meticulously craft prompt templates for each LLM in the pipeline, employing a few-shot learning approach. This includes presenting a set of example input-output pairs to guide the model in various tasks. All prompts used in this study are provided in Appendix \ref{app:prompts}. For the text-to-image generation task, we use Stable Diffusion 3.5 \citep{rombach2022highresolutionimagesynthesislatent} as the model for image synthesis.

\paragraph{Benchmarks} We evaluate the models on both culture-specific and culture-agnostic tasks. To assess cultural understanding, we adapt three benchmarks—GlobalRG-Grounding, GlobalRG-Retrieval \citep{bhatia2024local}, and CROPE \citep{nikandrou2024crope}—into statement-ranking tasks suitable for CLIP-based models. In each task, the model must select the most semantically accurate description for a given image from a set of culturally grounded statements, thereby testing its ability to capture fine-grained, culture-specific visual cues (reported as \textit{Accuracy}). To assess general vision-language capabilities, we further evaluate the models on MS COCO \citep{lin2015microsoftcococommonobjects} and Flickr30k \citep{plummer2016flickr30kentitiescollectingregiontophrase}, reporting the average \textit{Recall@5} for both image-to-text and text-to-image retrieval tasks. Details of the benchmarks and additional evaluation results on widely used image classification datasets are provided in Appendix~\ref{app:benchmark}.

\paragraph{Baseline} We first evaluate the performance of CLIP \citep{radford2021learning}, NegCLIP \citep{yuksekgonul2022and}, and TripletCLIP \citep{patel2024tripletclip} on our tasks to establish a baseline. Building on this, we introduce CLIP++, NegCLIP++, and TripletCLIP++ as enhanced baselines. In these ++ versions, we train the base CLIP model using our own dataset, aligning the synthetic images with contextualized caption-image pairs, without incorporating the concept.

\paragraph{Implementation Details} 
All fine-tuned CLIP models use ViT-B/32 \citep{dosovitskiy2021imageworth16x16words} as the image encoder \(\mathcal{F}\) and the default CLIP text encoder \citep{radford2021learning} as the text encoder \(\mathcal{G}\). We fine-tune the model using a parameter-efficient method, LoRA \citep{hu2021loralowrankadaptationlarge}, during which \(\mathcal{F}\) and \(\mathcal{G}\) are frozen, and additional LoRA parameters for these two encoders are applied and trained for 10 epochs with a global batch size of 2048, a learning rate of \(3 \times 10^{-6}\), weight decay of 0.1, and a cosine learning rate schedule. We employ LoRA primarily to maintain the model's general capability rather than to reduce memory requirements. In our preliminary experiments with full parameter fine-tuning, we observed significant performance degradation on both culture-specific and culture-agnostic benchmarks, likely due to the distribution gap between the general data used in pretraining and the culture-specific data used during fine-tuning, which disrupted knowledge the model had already acquired.  Before inference, LoRA parameters are merged with the backbone transformers, ensuring both efficiency and the preservation of zero-shot transfer capabilities. All models are fine-tuned on 4 Nvidia H20 GPUs using the official Hugging Face Transformers codebase \citep{wolf-etal-2020-transformers}.

\subsection{Main Results}

\paragraph{Evaluation on Culture-Specific Tasks}
As summarized in Table~\ref{tab:main results}, CultureCLIP significantly outperforms all baseline models on cultural benchmarks, achieving a 5.49\% improvement over CLIP on GlobalRG-G, thus demonstrating strong fine-grained cultural understanding. Directly fine-tuning CLIP on our cultural dataset (CLIP++) leads to a substantial performance drop of 17.93\%, highlighting the limitations of naive fine-tuning without explicit negative samples or concept-level alignment. In contrast, our enhanced variants NegCLIP++ and TripletCLIP++ (which incorporate hard negatives) achieve improvements of 2.34\% and 0.80\% over CLIP, respectively. When further combined with concept-level alignment in CultureCLIP, we observe a large net improvement, emphasizing the effectiveness of our design in leveraging hard negatives and concept supervision for cultural feature discrimination. Similar trends are observed for GlobalRG-R and CROPE. We note that NegCLIP and TripletCLIP are pre-trained based on much smaller datasets (e.g., CC3M, CC12M) and do not include culturally relevant data, resulting in substantially lower performance on general benchmarks. For fairness, we do not consider them direct baselines in the cultural evaluation but include them in the table for completeness. All models in Table~\ref{tab:main results} are trained on the same unfiltered 100k synthetic dataset using LoRA with rank 4 to ensure a fair comparison.

\paragraph{Evaluation on Culture-Agnostic Tasks}
On general vision-language tasks, CultureCLIP maintains strong performance and even slightly improves over the baseline, with gains of 0.90\% on MS COCO and 0.30\% on Flickr30k. This indicates that cultural fine-tuning does not compromise, and may even enhance, general retrieval capabilities.

\begin{table}[t]
    \centering
    \vspace{-1ex}
    \caption{Experimental results on culture-specific and culture-agnostic tasks. All models are trained on the same unfiltered 100k dataset using LoRA with rank 4. Best scores are in \textbf{bold}. Second best scores are underlined.}
    \resizebox{\linewidth}{!}{
    \begin{tabular}{lcc|ccc|cc}
        \toprule
        \multirow{2}{*}{\textbf{Methods}} & \multirow{2}{*}{\textbf{Neg}} & \multirow{2}{*}{\textbf{Con}} 
        & \multicolumn{3}{c|}{\textbf{Culture-Specific Tasks}} 
        & \multicolumn{2}{c}{\textbf{Culture-Agnostic Tasks}} \\
        \cmidrule(lr){4-6} \cmidrule(lr){7-8}
        & & & \textbf{GlobalRG-G} & \textbf{GlobalRG-R} & \textbf{CROPE} 
        & \textbf{MS COCO} & \textbf{Flickr30k} \\
        \midrule
        CLIP & $\times$ &$\times$ & 63.98 & 78.22 & 74.69 & 65.40 & 89.0 \\
        NegCLIP & $\checkmark$ &$\times$& - & - & - & 6.50 & 2.70 \\
        TripletCLIP & $\checkmark$ &$\times$& - & - & - & 10.80 & 22.00 \\
        \midrule
        CLIP++~(ours) & $\times$ &$\times$ & 46.05 & 49.98 & 73.62 & 28.80 & 50.80 \\
        NegCLIP++ (ours) & $\checkmark$ &$\times$ & \underline{66.32} & 78.41 & 79.25 & \underline{65.50} & \underline{89.20} \\
        TripletCLIP++ (ours) & $\checkmark$ &$\times$ & 64.78 & \underline{78.45} & \textbf{79.25} & \underline{65.50} & \textbf{89.30} \\
        \midrule
        \textbf{CultureCLIP (ours)} & $\checkmark$ & $\checkmark$ & \textbf{69.47} & \textbf{78.60} & \underline{78.84} & \textbf{66.30} & \textbf{89.30} \\
        \bottomrule
    \end{tabular}
    }
    \vspace{-2ex}
    \label{tab:main results}
\end{table}

\subsection{Ablations}

\paragraph{Which contributes more to the model and alignment, caption or concept?} As shown in Table~\ref{tab:ablation}, concepts act as abstract semantic anchors, providing stronger cultural alignment capacity and enabling the model to distinguish subtle differences. Captions refine the model's understanding of specific details but are less critical for recognizing cultural nuances.

\begin{table}[t]
    \centering
    \vspace{-2ex}
    \caption{
    Ablation study on loss configurations. All models are trained on the same unfiltered 100k dataset using LoRA (rank 4) to preserve general multimodal alignment capabilities.}
    \label{tab:ablation}
    \resizebox{\linewidth}{!}{
    \begin{tabular}{lc|ccc|cc}
    \toprule
    \multirow{2}{*}{\textbf{Configuration}} & \multirow{2}{*}{\boldsymbol{$\lambda$ }(cap/con)}
    & \multicolumn{3}{c|}{\textbf{Culture-Specific Tasks}} 
    & \multicolumn{2}{c}{\textbf{Culture-Agnostic Tasks}} \\
    \cmidrule(lr){3-5} \cmidrule(lr){6-7}
    & & \textbf{GlobalRG-G} & \textbf{GlobalRG-R} & \textbf{CROPE} 
    & \textbf{MS COCO} & \textbf{Flickr30k} \\
    \midrule
    \multicolumn{7}{l}{\textit{Single Branch (No Negative)}} \\
    Caption-only w/o neg & 1.0 / -- & 66.95 & 77.43 & 79.19 & 65.60 & 89.20 \\
    Concept-only w/o neg & -- / 1.0 & 64.24 & 77.70 & 79.19 & 65.50 & 89.00 \\
    \midrule
    \multicolumn{7}{l}{\textit{Single Branch (With Negative)}} \\
    Caption-only w/ neg & 1.0 / -- & 66.27 & 77.53 & \underline{79.25} & 65.50 & 89.20 \\
    Concept-only w/ neg & -- / 1.0 & 65.83 & 77.29 & 79.19 & 65.60 & 88.90 \\
    \midrule
    \multicolumn{7}{l}{\textit{Mixed Branches}} \\
    Cap (w/o neg) + Con (w/o neg) & 0.5 / 0.5 & 67.29 & 77.27 & \underline{79.25} & 65.50 & 89.20 \\
    Cap (w/ neg) + Con (w/o neg) & 0.5 / 0.5 & 67.12 & \underline{78.70} & \underline{79.25} & 65.60 & \underline{89.30} \\
    Cap (w/o neg) + Con (w/ neg) & 0.5 / 0.5 & \underline{68.81} & 76.87 & 79.19 & 65.50 & 89.20 \\
    \midrule
    \multicolumn{7}{l}{\textit{Full (Both with Negative)}} \\
    Both w/ neg (Ours) & 0.7 / 0.3 & 65.93 & \textbf{78.80} & \textbf{79.37} & \textbf{66.80} & \textbf{89.50} \\
    Both w/ neg (Ours) & 0.5 / 0.5 & 67.12 & 78.25 & 78.60 & \underline{66.30} & 89.20 \\
    Both w/ neg (Ours) & 0.3 / 0.7 & \textbf{69.47} & 78.60 & 78.84 & 66.10 & \underline{89.30} \\
    \bottomrule
    \end{tabular}
    }
\end{table}

\paragraph{Can a higher-quality cultural dataset improve performance?}
As shown in Table~\ref{tab:ablation_qc_lora}, using quality-filtered data (“+QF”), which contains 73.8k high-quality samples after image filtering, consistently improves performance despite using fewer samples compared to the full 100k unfiltered set. For example, comparing Config 5 (+QF, LoRA r=4) to Config 3 (LoRA r=4), and Config 6 (+QF, LoRA r=8) to Config 4 (LoRA r=8), we observe clear performance gains across all cultural benchmarks. This underscores the effectiveness of our data curation pipeline in providing cleaner and more informative supervision for fine-grained cultural alignment.

\begin{wraptable}{r}{0.65\textwidth}
    \centering
    \vspace{-3.5ex}
    \caption{Ablation study on quality filtering (QF) and LoRA. “+QF” uses the 73.8k filtered samples; otherwise, the full 100k dataset is used.}
    \label{tab:ablation_qc_lora}
    \resizebox{\linewidth}{!}{
    \begin{tabular}{lcc|ccc}
    \toprule
    \textbf{Configuration} & \textbf{QF} & \textbf{LoRA Rank} & \textbf{GlobalRG-G} & \textbf{GlobalRG-R} & \textbf{CROPE} \\
    \midrule
    Baseline & $\times$ & - & 46.95 & 49.98 & 73.62 \\
    + QF & \checkmark & - & 43.64 & 50.68 & 74.33 \\
    + LoRA (r=4) & $\times$ & 4 & \underline{69.47} & \textbf{78.60} & 78.84 \\
    + LoRA (r=8) & $\times$ & 8 & 65.29 & 78.03 & \textbf{79.37} \\
    + QF + LoRA (r=4) & \checkmark & 4 & \textbf{69.67} & \textbf{78.60} & \underline{79.21} \\
    + QF + LoRA (r=8) & \checkmark & 8 & 66.48 & 78.21 & \textbf{79.37} \\
    \bottomrule
    \vspace{-4ex}
    \end{tabular}
    }
\end{wraptable}
\paragraph{What role does LoRA play in adapting a pretrained model to downstream tasks?} Our ablation results in Table~\ref{tab:ablation_qc_lora} show that without LoRA (e.g., Configs 1 and 2), directly fine-tuning a pretrained model on a domain-specific cultural dataset leads to a substantial performance drop—specifically, 22.52\% and 26.03\% lower compared to Configs 3 and 5, respectively. This suggests that, without LoRA, the model struggles to effectively absorb specialized supervision, resulting in a significant loss of generalization ability. In contrast, when LoRA is applied (Configs 3–6), the model not only achieves improved performance on cultural benchmarks but also maintains its capabilities on general tasks. These findings highlight LoRA’s essential role in mitigating catastrophic forgetting: it enables the model to flexibly adapt to new cultural signals while preserving the broad vision-language alignment learned during pretraining, ensuring robustness across both domain-specific and general scenarios.
\section{Conclusion}

In this paper, we introduce CulTwin, a high-quality synthetic dataset of paired concept-caption-image triplets verified by humans, where captions are enriched with cultural background knowledge using a vision-language model, and images are generated by a text-to-image model to reflect fine-grained visual features. Building on CulTwin, we propose CultureCLIP, a novel contrastive learning framework that jointly aligns cultural concepts, captions, and images in a shared embedding space. Our experiments demonstrate that CultureCLIP surpasses baseline models on culture-specific benchmarks, achieving a 5.49\% improvement while simultaneously showing performance gains rather than degradation on culture-agnostic benchmarks. These results underscore the effectiveness of our synthetic dataset and training methodology in capturing nuanced cultural distinctions while preserving and even enhancing the model's generalization capabilities across broader contexts.
 
\section{Limitations and Future Work}

While CultureCLIP significantly improves fine-grained cultural understanding, several limitations remain. Both CLIP and CultureCLIP still struggle with cases where the visual distinction is highly abstract or stylistic (see error case analysis in Appendix~\ref{app:error cases}). In addition, our current pipeline, for practical considerations, adopts Qwen2.5-VL as an MLLM-as-a-Judge to assess cultural relevance through multidimensional scoring and to guide filtering. However, compared to more advanced models such as GPT-4o, this choice may introduce biases, lead to misjudgments, or lack interpretability. Furthermore, despite the diversity of CulTwin, it is fundamentally a fully synthetic dataset, and there may still exist a distributional gap between synthetic and real-world images. Looking ahead, future work may explore several directions: \textbf{(1) Improving abstract visual reasoning}, by enhancing the model’s capacity to recognize subtle visual cues, such as artistic styles or symbolic meanings; \textbf{(2) Enhancing the robustness of cultural understanding}, by mitigating vulnerability to visual variations and ensuring stable performance across diverse conditions \citep{fan2025unveilinglacklvlmrobustness}; \textbf{(3) Developing interpretable assessment modules}, to enable more robust and transparent cultural data evaluation beyond the current Qwen2.5-VL-based judge; and \textbf{(4) Bridging the synthetic-real domain gap}, by integrating real and synthetic data or applying domain adaptation strategies to improve generalization and visual grounding.

\newpage
\bibliography{colm2025_conference}

\begin{thebibliography}{42}
\providecommand{\natexlab}[1]{#1}
\providecommand{\url}[1]{\texttt{#1}}
\expandafter\ifx\csname urlstyle\endcsname\relax
  \providecommand{\doi}[1]{doi: #1}\else
  \providecommand{\doi}{doi: \begingroup \urlstyle{rm}\Url}\fi

\bibitem[Bai et~al.(2025)Bai, Chen, Liu, Wang, Ge, Song, Dang, Wang, Wang, Tang, Zhong, Zhu, Yang, Li, Wan, Wang, Ding, Fu, Xu, Ye, Zhang, Xie, Cheng, Zhang, Yang, Xu, and Lin]{bai2025qwen25vltechnicalreport}
Shuai Bai, Keqin Chen, Xuejing Liu, Jialin Wang, Wenbin Ge, Sibo Song, Kai Dang, Peng Wang, Shijie Wang, Jun Tang, Humen Zhong, Yuanzhi Zhu, Mingkun Yang, Zhaohai Li, Jianqiang Wan, Pengfei Wang, Wei Ding, Zheren Fu, Yiheng Xu, Jiabo Ye, Xi~Zhang, Tianbao Xie, Zesen Cheng, Hang Zhang, Zhibo Yang, Haiyang Xu, and Junyang Lin.
\newblock Qwen2.5-vl technical report, 2025.
\newblock URL \url{https://arxiv.org/abs/2502.13923}.

\bibitem[Baldrati et~al.(2023)Baldrati, Bertini, Uricchio, and del Bimbo]{baldrati2023composedimageretrievalusing}
Alberto Baldrati, Marco Bertini, Tiberio Uricchio, and Alberto del Bimbo.
\newblock Composed image retrieval using contrastive learning and task-oriented clip-based features, 2023.
\newblock URL \url{https://arxiv.org/abs/2308.11485}.

\bibitem[Bhatia et~al.(2024)Bhatia, Ravi, Chinchure, Hwang, and Shwartz]{bhatia2024local}
Mehar Bhatia, Sahithya Ravi, Aditya Chinchure, Eunjeong Hwang, and Vered Shwartz.
\newblock From local concepts to universals: Evaluating the multicultural understanding of vision-language models.
\newblock \emph{arXiv preprint arXiv:2407.00263}, 2024.

\bibitem[Brown et~al.(2020)Brown, Mann, Ryder, Subbiah, Kaplan, Dhariwal, Neelakantan, Shyam, Sastry, Askell, Agarwal, Herbert-Voss, Krueger, Henighan, Child, Ramesh, Ziegler, Wu, Winter, Hesse, Chen, Sigler, Litwin, Gray, Chess, Clark, Berner, McCandlish, Radford, Sutskever, and Amodei]{brown2020languagemodelsfewshotlearners}
Tom~B. Brown, Benjamin Mann, Nick Ryder, Melanie Subbiah, Jared Kaplan, Prafulla Dhariwal, Arvind Neelakantan, Pranav Shyam, Girish Sastry, Amanda Askell, Sandhini Agarwal, Ariel Herbert-Voss, Gretchen Krueger, Tom Henighan, Rewon Child, Aditya Ramesh, Daniel~M. Ziegler, Jeffrey Wu, Clemens Winter, Christopher Hesse, Mark Chen, Eric Sigler, Mateusz Litwin, Scott Gray, Benjamin Chess, Jack Clark, Christopher Berner, Sam McCandlish, Alec Radford, Ilya Sutskever, and Dario Amodei.
\newblock Language models are few-shot learners, 2020.
\newblock URL \url{https://arxiv.org/abs/2005.14165}.

\bibitem[Chen et~al.(2024)Chen, Chen, Zhang, Liu, Wang, Zhou, Zhang, Wan, Zhou, and Sun]{chen2024mllmasajudgeassessingmultimodalllmasajudge}
Dongping Chen, Ruoxi Chen, Shilin Zhang, Yinuo Liu, Yaochen Wang, Huichi Zhou, Qihui Zhang, Yao Wan, Pan Zhou, and Lichao Sun.
\newblock Mllm-as-a-judge: Assessing multimodal llm-as-a-judge with vision-language benchmark, 2024.
\newblock URL \url{https://arxiv.org/abs/2402.04788}.

\bibitem[Dosovitskiy et~al.(2021)Dosovitskiy, Beyer, Kolesnikov, Weissenborn, Zhai, Unterthiner, Dehghani, Minderer, Heigold, Gelly, Uszkoreit, and Houlsby]{dosovitskiy2021imageworth16x16words}
Alexey Dosovitskiy, Lucas Beyer, Alexander Kolesnikov, Dirk Weissenborn, Xiaohua Zhai, Thomas Unterthiner, Mostafa Dehghani, Matthias Minderer, Georg Heigold, Sylvain Gelly, Jakob Uszkoreit, and Neil Houlsby.
\newblock An image is worth 16x16 words: Transformers for image recognition at scale, 2021.
\newblock URL \url{https://arxiv.org/abs/2010.11929}.

\bibitem[Fan et~al.(2023)Fan, Krishnan, Isola, Katabi, and Tian]{fan2023improving}
Lijie Fan, Dilip Krishnan, Phillip Isola, Dina Katabi, and Yonglong Tian.
\newblock Improving clip training with language rewrites.
\newblock \emph{Advances in Neural Information Processing Systems}, 36:\penalty0 35544--35575, 2023.

\bibitem[Fan et~al.(2025)Fan, Wang, Polisetty, and Fung]{fan2025unveilinglacklvlmrobustness}
Zhiyuan Fan, Yumeng Wang, Sandeep Polisetty, and Yi~R. Fung.
\newblock Unveiling the lack of lvlm robustness to fundamental visual variations: Why and path forward, 2025.
\newblock URL \url{https://arxiv.org/abs/2504.16727}.

\bibitem[Fang et~al.(2022)Fang, Ilharco, Wortsman, Wan, Shankar, Dave, and Schmidt]{fang2022datadeterminesdistributionalrobustness}
Alex Fang, Gabriel Ilharco, Mitchell Wortsman, Yuhao Wan, Vaishaal Shankar, Achal Dave, and Ludwig Schmidt.
\newblock Data determines distributional robustness in contrastive language image pre-training (clip), 2022.
\newblock URL \url{https://arxiv.org/abs/2205.01397}.

\bibitem[Hammoud et~al.(2024)Hammoud, Itani, Pizzati, Torr, Bibi, and Ghanem]{hammoud2024synthclip}
Hasan Abed Al~Kader Hammoud, Hani Itani, Fabio Pizzati, Philip Torr, Adel Bibi, and Bernard Ghanem.
\newblock Synthclip: Are we ready for a fully synthetic clip training?
\newblock \emph{arXiv preprint arXiv:2402.01832}, 2024.

\bibitem[He et~al.(2025)He, Polisetty, Fan, Huang, Wu, and Fung]{he2025mmboundaryadvancingmllmknowledge}
Zhitao He, Sandeep Polisetty, Zhiyuan Fan, Yuchen Huang, Shujin Wu, and Yi~R. Fung.
\newblock Mmboundary: Advancing mllm knowledge boundary awareness through reasoning step confidence calibration, 2025.
\newblock URL \url{https://arxiv.org/abs/2505.23224}.

\bibitem[Hu et~al.(2021)Hu, Shen, Wallis, Allen-Zhu, Li, Wang, Wang, and Chen]{hu2021loralowrankadaptationlarge}
Edward~J. Hu, Yelong Shen, Phillip Wallis, Zeyuan Allen-Zhu, Yuanzhi Li, Shean Wang, Lu~Wang, and Weizhu Chen.
\newblock Lora: Low-rank adaptation of large language models, 2021.
\newblock URL \url{https://arxiv.org/abs/2106.09685}.

\bibitem[Jia et~al.(2021)Jia, Yang, Xia, Chen, Parekh, Pham, Le, Sung, Li, and Duerig]{jia2021scalingvisualvisionlanguagerepresentation}
Chao Jia, Yinfei Yang, Ye~Xia, Yi-Ting Chen, Zarana Parekh, Hieu Pham, Quoc~V. Le, Yunhsuan Sung, Zhen Li, and Tom Duerig.
\newblock Scaling up visual and vision-language representation learning with noisy text supervision, 2021.
\newblock URL \url{https://arxiv.org/abs/2102.05918}.

\bibitem[Koukounas et~al.(2024)Koukounas, Mastrapas, Günther, Wang, Martens, Mohr, Sturua, Akram, Martínez, Ognawala, Guzman, Werk, Wang, and Xiao]{koukounas2024jinaclipclipmodel}
Andreas Koukounas, Georgios Mastrapas, Michael Günther, Bo~Wang, Scott Martens, Isabelle Mohr, Saba Sturua, Mohammad~Kalim Akram, Joan~Fontanals Martínez, Saahil Ognawala, Susana Guzman, Maximilian Werk, Nan Wang, and Han Xiao.
\newblock Jina clip: Your clip model is also your text retriever, 2024.
\newblock URL \url{https://arxiv.org/abs/2405.20204}.

\bibitem[Li et~al.(2023)Li, Li, Savarese, and Hoi]{li2023blip}
Junnan Li, Dongxu Li, Silvio Savarese, and Steven Hoi.
\newblock Blip-2: Bootstrapping language-image pre-training with frozen image encoders and large language models.
\newblock In \emph{International conference on machine learning}, pp.\  19730--19742. PMLR, 2023.

\bibitem[Li et~al.(2024)Li, Zhang, Wang, Hao, Lei, Tan, Zhou, Liu, Yang, Xiong, et~al.]{li2024chemvlm}
Junxian Li, Di~Zhang, Xunzhi Wang, Zeying Hao, Jingdi Lei, Qian Tan, Cai Zhou, Wei Liu, Yaotian Yang, Xinrui Xiong, et~al.
\newblock Chemvlm: Exploring the power of multimodal large language models in chemistry area.
\newblock \emph{arXiv preprint arXiv:2408.07246}, 2024.

\bibitem[Lin et~al.(2015)Lin, Maire, Belongie, Bourdev, Girshick, Hays, Perona, Ramanan, Zitnick, and Dollár]{lin2015microsoftcococommonobjects}
Tsung-Yi Lin, Michael Maire, Serge Belongie, Lubomir Bourdev, Ross Girshick, James Hays, Pietro Perona, Deva Ramanan, C.~Lawrence Zitnick, and Piotr Dollár.
\newblock Microsoft coco: Common objects in context, 2015.
\newblock URL \url{https://arxiv.org/abs/1405.0312}.

\bibitem[Liu et~al.(2025)Liu, Jin, Li, Wong, Wen, Sun, Chen, Xie, and Wang]{liu2025culturevlm}
Shudong Liu, Yiqiao Jin, Cheng Li, Derek~F Wong, Qingsong Wen, Lichao Sun, Haipeng Chen, Xing Xie, and Jindong Wang.
\newblock Culturevlm: Characterizing and improving cultural understanding of vision-language models for over 100 countries.
\newblock \emph{arXiv preprint arXiv:2501.01282}, 2025.

\bibitem[Lu et~al.(2024)Lu, Liu, Zhang, Wang, Dong, Liu, Sun, Ren, Li, Yang, et~al.]{lu2024deepseek}
Haoyu Lu, Wen Liu, Bo~Zhang, Bingxuan Wang, Kai Dong, Bo~Liu, Jingxiang Sun, Tongzheng Ren, Zhuoshu Li, Hao Yang, et~al.
\newblock Deepseek-vl: towards real-world vision-language understanding.
\newblock \emph{arXiv preprint arXiv:2403.05525}, 2024.

\bibitem[Moor et~al.(2023)Moor, Huang, Wu, Yasunaga, Dalmia, Leskovec, Zakka, Reis, and Rajpurkar]{moor2023med}
Michael Moor, Qian Huang, Shirley Wu, Michihiro Yasunaga, Yash Dalmia, Jure Leskovec, Cyril Zakka, Eduardo~Pontes Reis, and Pranav Rajpurkar.
\newblock Med-flamingo: a multimodal medical few-shot learner.
\newblock In \emph{Machine Learning for Health (ML4H)}, pp.\  353--367. PMLR, 2023.

\bibitem[Nayak et~al.(2024)Nayak, Jain, Awal, Reddy, Van~Steenkiste, Hendricks, Agrawal, et~al.]{nayak2024benchmarking}
Shravan Nayak, Kanishk Jain, Rabiul Awal, Siva Reddy, Sjoerd Van~Steenkiste, Lisa~Anne Hendricks, Aishwarya Agrawal, et~al.
\newblock Benchmarking vision language models for cultural understanding.
\newblock \emph{arXiv preprint arXiv:2407.10920}, 2024.

\bibitem[Nguyen et~al.(2023)Nguyen, Ilharco, Wortsman, Oh, and Schmidt]{nguyen2023qualityquantityinteractiondataset}
Thao Nguyen, Gabriel Ilharco, Mitchell Wortsman, Sewoong Oh, and Ludwig Schmidt.
\newblock Quality not quantity: On the interaction between dataset design and robustness of clip, 2023.
\newblock URL \url{https://arxiv.org/abs/2208.05516}.

\bibitem[Nikandrou et~al.(2024)Nikandrou, Pantazopoulos, Vitsakis, Konstas, and Suglia]{nikandrou2024crope}
Malvina Nikandrou, Georgios Pantazopoulos, Nikolas Vitsakis, Ioannis Konstas, and Alessandro Suglia.
\newblock Crope: Evaluating in-context adaptation of vision and language models to culture-specific concepts.
\newblock \emph{arXiv preprint arXiv:2410.15453}, 2024.

\bibitem[Patel et~al.(2024)Patel, Kusumba, Cheng, Kim, Gokhale, Baral, et~al.]{patel2024tripletclip}
Maitreya Patel, Naga Sai~Abhiram Kusumba, Sheng Cheng, Changhoon Kim, Tejas Gokhale, Chitta Baral, et~al.
\newblock Tripletclip: Improving compositional reasoning of clip via synthetic vision-language negatives.
\newblock \emph{Advances in Neural Information Processing Systems}, 37:\penalty0 32731--32760, 2024.

\bibitem[Peng et~al.(2025)Peng, Zhang, Zhang, You, Liu, Zhu, Yang, Xu, Geng, and Yang]{peng2025lmmr1empowering3blmms}
Yingzhe Peng, Gongrui Zhang, Miaosen Zhang, Zhiyuan You, Jie Liu, Qipeng Zhu, Kai Yang, Xingzhong Xu, Xin Geng, and Xu~Yang.
\newblock Lmm-r1: Empowering 3b lmms with strong reasoning abilities through two-stage rule-based rl, 2025.
\newblock URL \url{https://arxiv.org/abs/2503.07536}.

\bibitem[Plummer et~al.(2016)Plummer, Wang, Cervantes, Caicedo, Hockenmaier, and Lazebnik]{plummer2016flickr30kentitiescollectingregiontophrase}
Bryan~A. Plummer, Liwei Wang, Chris~M. Cervantes, Juan~C. Caicedo, Julia Hockenmaier, and Svetlana Lazebnik.
\newblock Flickr30k entities: Collecting region-to-phrase correspondences for richer image-to-sentence models, 2016.
\newblock URL \url{https://arxiv.org/abs/1505.04870}.

\bibitem[Radford et~al.(2021)Radford, Kim, Hallacy, Ramesh, Goh, Agarwal, Sastry, Askell, Mishkin, Clark, et~al.]{radford2021learning}
Alec Radford, Jong~Wook Kim, Chris Hallacy, Aditya Ramesh, Gabriel Goh, Sandhini Agarwal, Girish Sastry, Amanda Askell, Pamela Mishkin, Jack Clark, et~al.
\newblock Learning transferable visual models from natural language supervision.
\newblock In \emph{International conference on machine learning}, pp.\  8748--8763. PmLR, 2021.

\bibitem[Rombach et~al.(2022)Rombach, Blattmann, Lorenz, Esser, and Ommer]{rombach2022highresolutionimagesynthesislatent}
Robin Rombach, Andreas Blattmann, Dominik Lorenz, Patrick Esser, and Björn Ommer.
\newblock High-resolution image synthesis with latent diffusion models, 2022.
\newblock URL \url{https://arxiv.org/abs/2112.10752}.

\bibitem[Romero et~al.(2024)Romero, Lyu, Wibowo, Lynn, Hamed, Kishore, Mandal, Dragonetti, Abzaliev, Tonja, Balcha, Whitehouse, Salamea, Velasco, Adelani, Meur, Villa-Cueva, Koto, Farooqui, Belcavello, Batnasan, Vallejo, Caulfield, Ivetta, Song, Ademtew, Maina, Lovenia, Azime, Cruz, Gala, Geng, Ortiz-Barajas, Baek, Dunstan, Alemany, Nagasinghe, Benotti, D'Haro, Viridiano, Estecha-Garitagoitia, Cabrera, Rodríguez-Cantelar, Jouitteau, Mihaylov, Imam, Adilazuarda, Gochoo, Otgonbold, Etori, Niyomugisha, Silva, Chitale, Dabre, Chevi, Zhang, Diandaru, Cahyawijaya, Góngora, Jeong, Purkayastha, Kuribayashi, Clifford, Jayakumar, Torrent, Ehsan, Araujo, Kementchedjhieva, Burzo, Lim, Yong, Ignat, Nwatu, Mihalcea, Solorio, and Aji]{romero2024cvqaculturallydiversemultilingualvisual}
David Romero, Chenyang Lyu, Haryo~Akbarianto Wibowo, Teresa Lynn, Injy Hamed, Aditya~Nanda Kishore, Aishik Mandal, Alina Dragonetti, Artem Abzaliev, Atnafu~Lambebo Tonja, Bontu~Fufa Balcha, Chenxi Whitehouse, Christian Salamea, Dan~John Velasco, David~Ifeoluwa Adelani, David~Le Meur, Emilio Villa-Cueva, Fajri Koto, Fauzan Farooqui, Frederico Belcavello, Ganzorig Batnasan, Gisela Vallejo, Grainne Caulfield, Guido Ivetta, Haiyue Song, Henok~Biadglign Ademtew, Hernán Maina, Holy Lovenia, Israel~Abebe Azime, Jan Christian~Blaise Cruz, Jay Gala, Jiahui Geng, Jesus-German Ortiz-Barajas, Jinheon Baek, Jocelyn Dunstan, Laura~Alonso Alemany, Kumaranage Ravindu~Yasas Nagasinghe, Luciana Benotti, Luis~Fernando D'Haro, Marcelo Viridiano, Marcos Estecha-Garitagoitia, Maria Camila~Buitrago Cabrera, Mario Rodríguez-Cantelar, Mélanie Jouitteau, Mihail Mihaylov, Mohamed Fazli~Mohamed Imam, Muhammad~Farid Adilazuarda, Munkhjargal Gochoo, Munkh-Erdene Otgonbold, Naome Etori, Olivier Niyomugisha, Paula~Mónica Silva, Pranjal
  Chitale, Raj Dabre, Rendi Chevi, Ruochen Zhang, Ryandito Diandaru, Samuel Cahyawijaya, Santiago Góngora, Soyeong Jeong, Sukannya Purkayastha, Tatsuki Kuribayashi, Teresa Clifford, Thanmay Jayakumar, Tiago~Timponi Torrent, Toqeer Ehsan, Vladimir Araujo, Yova Kementchedjhieva, Zara Burzo, Zheng~Wei Lim, Zheng~Xin Yong, Oana Ignat, Joan Nwatu, Rada Mihalcea, Thamar Solorio, and Alham~Fikri Aji.
\newblock Cvqa: Culturally-diverse multilingual visual question answering benchmark, 2024.
\newblock URL \url{https://arxiv.org/abs/2406.05967}.

\bibitem[Saha et~al.(2024)Saha, Horn, and Maji]{saha2024improvedzeroshotclassificationadapting}
Oindrila Saha, Grant~Van Horn, and Subhransu Maji.
\newblock Improved zero-shot classification by adapting vlms with text descriptions, 2024.
\newblock URL \url{https://arxiv.org/abs/2401.02460}.

\bibitem[Schuhmann et~al.(2022)Schuhmann, Beaumont, Vencu, Gordon, Wightman, Cherti, Coombes, Katta, Mullis, Wortsman, et~al.]{schuhmann2022laion}
Christoph Schuhmann, Romain Beaumont, Richard Vencu, Cade Gordon, Ross Wightman, Mehdi Cherti, Theo Coombes, Aarush Katta, Clayton Mullis, Mitchell Wortsman, et~al.
\newblock Laion-5b: An open large-scale dataset for training next generation image-text models.
\newblock \emph{Advances in neural information processing systems}, 35:\penalty0 25278--25294, 2022.

\bibitem[Shen et~al.(2021)Shen, Li, Tan, Bansal, Rohrbach, Chang, Yao, and Keutzer]{shen2021clipbenefitvisionandlanguagetasks}
Sheng Shen, Liunian~Harold Li, Hao Tan, Mohit Bansal, Anna Rohrbach, Kai-Wei Chang, Zhewei Yao, and Kurt Keutzer.
\newblock How much can clip benefit vision-and-language tasks?, 2021.
\newblock URL \url{https://arxiv.org/abs/2107.06383}.

\bibitem[Song et~al.(2022)Song, Dong, Zhang, Liu, and Wei]{song-etal-2022-clip}
Haoyu Song, Li~Dong, Weinan Zhang, Ting Liu, and Furu Wei.
\newblock {CLIP} models are few-shot learners: Empirical studies on {VQA} and visual entailment.
\newblock In Smaranda Muresan, Preslav Nakov, and Aline Villavicencio (eds.), \emph{Proceedings of the 60th Annual Meeting of the Association for Computational Linguistics (Volume 1: Long Papers)}, pp.\  6088--6100, Dublin, Ireland, May 2022. Association for Computational Linguistics.
\newblock \doi{10.18653/v1/2022.acl-long.421}.
\newblock URL \url{https://aclanthology.org/2022.acl-long.421/}.

\bibitem[Wolf et~al.(2020)Wolf, Debut, Sanh, Chaumond, Delangue, Moi, Cistac, Rault, Louf, Funtowicz, Davison, Shleifer, von Platen, Ma, Jernite, Plu, Xu, Scao, Gugger, Drame, Lhoest, and Rush]{wolf-etal-2020-transformers}
Thomas Wolf, Lysandre Debut, Victor Sanh, Julien Chaumond, Clement Delangue, Anthony Moi, Pierric Cistac, Tim Rault, Rémi Louf, Morgan Funtowicz, Joe Davison, Sam Shleifer, Patrick von Platen, Clara Ma, Yacine Jernite, Julien Plu, Canwen Xu, Teven~Le Scao, Sylvain Gugger, Mariama Drame, Quentin Lhoest, and Alexander~M. Rush.
\newblock Transformers: State-of-the-art natural language processing.
\newblock In \emph{Proceedings of the 2020 Conference on Empirical Methods in Natural Language Processing: System Demonstrations}, pp.\  38--45, Online, October 2020. Association for Computational Linguistics.
\newblock URL \url{https://www.aclweb.org/anthology/2020.emnlp-demos.6}.

\bibitem[Xu et~al.(2023)Xu, Xie, Tan, Huang, Howes, Sharma, Li, Ghosh, Zettlemoyer, and Feichtenhofer]{xu2023demystifying}
Hu~Xu, Saining Xie, Xiaoqing~Ellen Tan, Po-Yao Huang, Russell Howes, Vasu Sharma, Shang-Wen Li, Gargi Ghosh, Luke Zettlemoyer, and Christoph Feichtenhofer.
\newblock Demystifying clip data.
\newblock \emph{arXiv preprint arXiv:2309.16671}, 2023.

\bibitem[Yuksekgonul et~al.(2022)Yuksekgonul, Bianchi, Kalluri, Jurafsky, and Zou]{yuksekgonul2022and}
Mert Yuksekgonul, Federico Bianchi, Pratyusha Kalluri, Dan Jurafsky, and James Zou.
\newblock When and why vision-language models behave like bags-of-words, and what to do about it?
\newblock \emph{arXiv preprint arXiv:2210.01936}, 2022.

\bibitem[Zhang et~al.(2024{\natexlab{a}})Zhang, Zhang, Dong, Zang, and Wang]{zhang2024long}
Beichen Zhang, Pan Zhang, Xiaoyi Dong, Yuhang Zang, and Jiaqi Wang.
\newblock Long-clip: Unlocking the long-text capability of clip.
\newblock In \emph{European Conference on Computer Vision}, pp.\  310--325. Springer, 2024{\natexlab{a}}.

\bibitem[Zhang et~al.(2024{\natexlab{b}})Zhang, Zhang, Zhang, and Tresp]{zhang2024can}
Gengyuan Zhang, Yurui Zhang, Kerui Zhang, and Volker Tresp.
\newblock Can vision-language models be a good guesser? exploring vlms for times and location reasoning.
\newblock In \emph{Proceedings of the IEEE/CVF Winter Conference on Applications of Computer Vision}, pp.\  636--645, 2024{\natexlab{b}}.

\bibitem[Zhang et~al.(2025)Zhang, Yao, Pi, Liang, and Fung]{zhang2025vlm2benchcloserlookvlms}
Jianshu Zhang, Dongyu Yao, Renjie Pi, Paul~Pu Liang, and Yi~R. Fung.
\newblock Vlm2-bench: A closer look at how well vlms implicitly link explicit matching visual cues, 2025.
\newblock URL \url{https://arxiv.org/abs/2502.12084}.

\bibitem[Zhang et~al.(2024{\natexlab{c}})Zhang, Awal, and Agrawal]{zhang2024contrasting}
Le~Zhang, Rabiul Awal, and Aishwarya Agrawal.
\newblock Contrasting intra-modal and ranking cross-modal hard negatives to enhance visio-linguistic compositional understanding.
\newblock In \emph{Proceedings of the IEEE/CVF Conference on Computer Vision and Pattern Recognition}, pp.\  13774--13784, 2024{\natexlab{c}}.

\bibitem[Zhao et~al.(2025)Zhao, Wei, and Bo]{zhao2025r1omniexplainableomnimultimodalemotion}
Jiaxing Zhao, Xihan Wei, and Liefeng Bo.
\newblock R1-omni: Explainable omni-multimodal emotion recognition with reinforcement learning, 2025.
\newblock URL \url{https://arxiv.org/abs/2503.05379}.

\bibitem[Zhou et~al.(2022)Zhou, Yang, Loy, and Liu]{Zhou_2022}
Kaiyang Zhou, Jingkang Yang, Chen~Change Loy, and Ziwei Liu.
\newblock Learning to prompt for vision-language models.
\newblock \emph{International Journal of Computer Vision}, 130\penalty0 (9):\penalty0 2337–2348, July 2022.
\newblock ISSN 1573-1405.
\newblock \doi{10.1007/s11263-022-01653-1}.
\newblock URL \url{http://dx.doi.org/10.1007/s11263-022-01653-1}.

\end{thebibliography}
\bibliographystyle{colm2025_conference}

\newpage
\clearpage
\appendix

\section{Country List and Cultural Taxonomy}
\label{app:taxonomy}
In this study, the list of country names is based on data from the GeoNames database: \href{https://www.geonames.org/countries/}{GeoNames.org}. 

The cultural taxonomy includes the following categories, each representing a significant aspect of cultural identity:

\begin{itemize}
    \item \textbf{Cuisine}: Refers to the foods, culinary practices, and cooking methods that are unique to specific regions or cultures. This includes iconic dishes, preparation techniques, and the cultural background behind eating habits, as well as the importance of food in social and religious practices.
    
    \item \textbf{Clothing}: Encompasses traditional garments, accessories, and adornments from various cultures. It includes not only clothing but also items like jewelry, headwear, and footwear that hold cultural significance, reflecting identity, status, and traditions.
    
    \item \textbf{Animal \& Plants}: Describes the native species, both fauna and flora, that hold cultural importance. This category includes the use of animals and plants in mythology, cuisine, traditional medicine, and environmental practices, as well as their roles in folklore and symbolism.
    
    \item \textbf{Art}: Includes visual arts, sculptures, and other forms of artistic expression that represent a culture’s aesthetic and artistic heritage. This encompasses paintings, sculptures, performance arts, and crafts that reflect the identity, beliefs, and historical evolution of a community.
    
    \item \textbf{Architecture}: Refers to the design, style, and structures built by a particular culture. This includes traditional houses, temples, monuments, and public buildings that showcase the engineering, material use, and aesthetic values of the culture.
    
    \item \textbf{Daily Life}: Covers the everyday activities, routines, and practices that define how people in a particular culture live. This includes family roles, work habits, and leisure activities, as well as practices around health, education, and community.
    
    \item \textbf{Symbol}: Involves the symbols, logos, and imagery that carry cultural meaning. This category includes national flags, religious icons, mythological figures, and colors that convey beliefs, values, and identity in various contexts.
    
    \item \textbf{Festival}: Encompasses cultural festivals, holidays, and ceremonies, along with the associated customs, rituals, and practices. Examples include events like Chinese New Year, Diwali, and Christmas, each rich in traditions, foods, and rituals that symbolize community and heritage.
\end{itemize}

\newpage
\section{CulTwin Details}
\label{app:cultwin details}

\begin{figure*}[h]
  \centering
  \includegraphics[width=\textwidth]{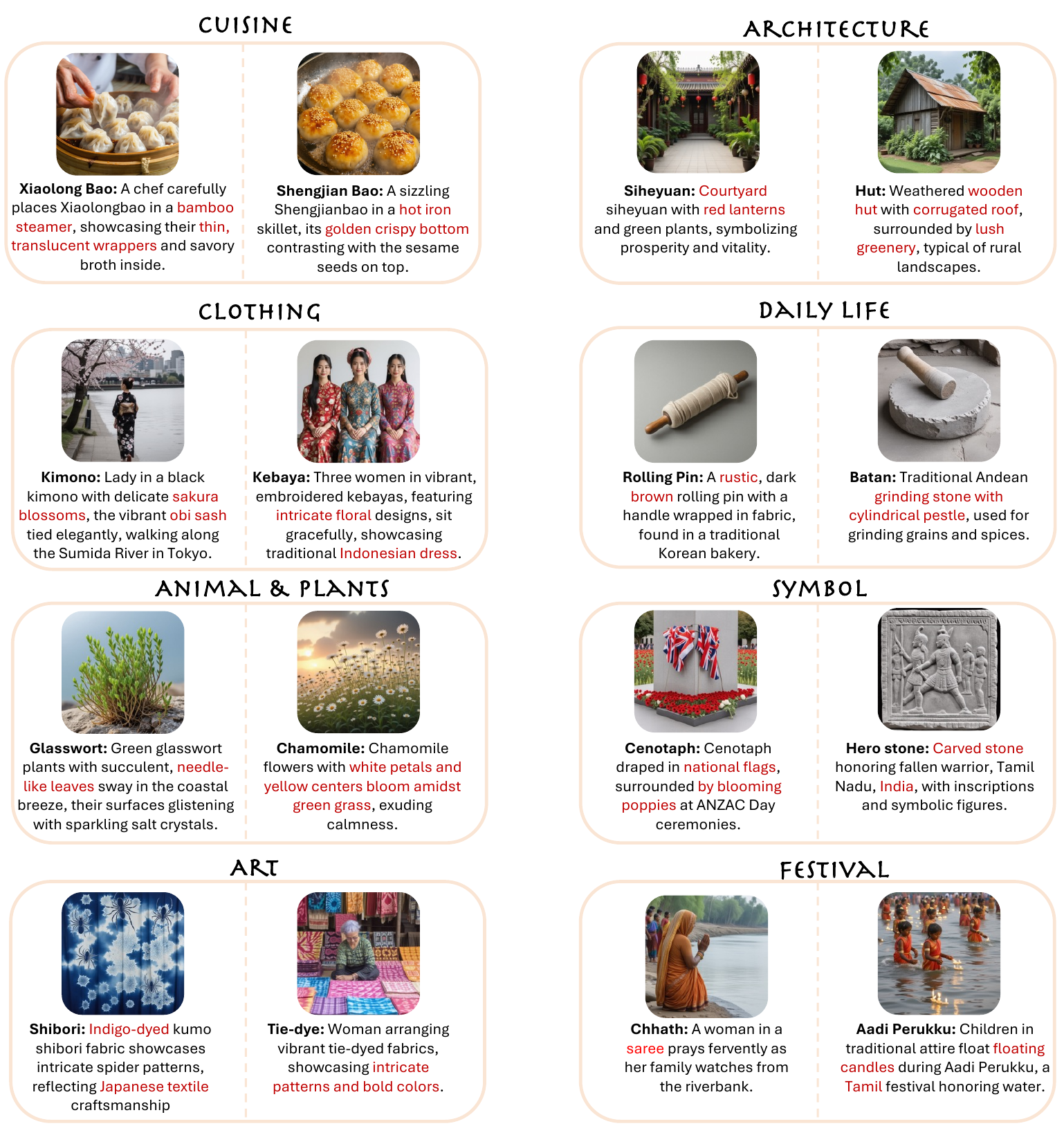}
  \vspace{-0.8em}
  \caption{Additional \textit{Twin Card} examples showcasing diverse cultural concepts beyond the main text examples, further illustrating fine-grained cultural distinctions and visual similarity.}
  \label{fig:more_examples}
\end{figure*}

We collected a total of 99,996 \textit{Twin Cards}, each consisting of two concept–caption–image triplets with culturally distinct negatives. After applying the image quality filtering described in Section~\ref{sec:stage3}, 73,823 high-quality samples were retained—a scale significantly larger than existing cultural benchmarks such as CROPE (approximately 1k samples). Each concept is represented as a single word, and the corresponding captions have an average length of 14.55 words. All images are synthetically generated from these captions using Stable Diffusion 3.5 Large Turbo, with an efficient generation throughput of approximately 3,000 images per hour on a single H20 GPU. Additional examples of \textit{Twin Cards} are illustrated in Figure~\ref{fig:more_examples}.

\section{CultureCLIP Pseudocode}
\begin{tcolorbox}[colframe=black, colback=white, coltitle=black, 
    boxrule=1mm, width=\textwidth, arc=4mm, auto outer arc]
\begin{lstlisting}[style=mystyle]
def cultureclip_loss(pos_concept_embs, pos_caption_embs, pos_img_embs,
                      neg_concept_embs, neg_caption_embs, neg_img_embs,
                      logit_scale, lambda_caption=0.5, lambda_concept=0.5):
    """
    Full CultureCLIP Loss: Caption's negclip + Concept's negclip
    Aligns concepts with their corresponding captions and images,
    while distinguishing them from their culturally opposite counterparts.
    """

    # Calculate Caption's NegCLIP loss
    caption_loss_val = (
        negclip_loss(pos_img_embs, pos_caption_embs, neg_caption_embs, logit_scale) + 
        negclip_loss(neg_img_embs, neg_caption_embs, pos_caption_embs, logit_scale)
    )
    
    # Calculate Concept's NegCLIP loss
    concept_loss_val = (
        negclip_loss(pos_img_embs, pos_concept_embs, neg_concept_embs, logit_scale) + 
        negclip_loss(neg_img_embs, neg_concept_embs, pos_concept_embs, logit_scale)
    )
    
    # Total Loss
    total_loss = lambda_caption * caption_loss_val + lambda_concept * concept_loss_val
    
    return total_loss
\end{lstlisting}
\end{tcolorbox}

\section{Benchmark Details and Additional Results}
\label{app:benchmark}

In this section, we provide detailed descriptions of the benchmarks used to evaluate our models, covering both culture-specific and culture-agnostic tasks.
\paragraph{Culture-Specific Tasks}
\begin{itemize}
    \item \textbf{GlobalRG-Grounding} \citep{bhatia2024local}: Each data point consists of an image, a concept, and a country. We generate four statement-based options for the model to choose from, such as "The item in the picture is \{concept\} in \{country\}." The correct option corresponds to the appropriate concept for that country, while incorrect options are created by selecting a concept from the same country that does not match the image.
    
    \item \textbf{GlobalRG-Retrieval} \citep{bhatia2024local}: Each data point consists of an image, a category, and a country, without specifying a particular concept. The task focuses on identifying the correct country for the depicted category. Options are phrased as "The picture depicts a kind of \{category\} in \{country\}," with incorrect options generated by randomly substituting the country.
    
    \item \textbf{CROPE} \citep{nikandrou2024crope}: The original dataset asks whether the image displays the defined concept ("yes" or "no"). We filter out "no" cases where the question concept and definition concept differ, and reformulate the task as a two-choice classification: "There is \{question concept\} in the image" or "There is \{definition concept\} in the image," requiring the model to correctly identify the depicted concept.
\end{itemize}

\paragraph{Culture-Agnostic Tasks}
\begin{itemize}
    \item \textbf{MS COCO} \citep{lin2015microsoftcococommonobjects}: This large-scale dataset comprises over 330,000 images annotated with approximately five captions each. We evaluate using bidirectional retrieval metrics, specifically \textit{Text2Image} and \textit{Image2Text} Recall@5, and report their arithmetic mean as the final score.
    
    \item \textbf{Flickr30k} \citep{plummer2016flickr30kentitiescollectingregiontophrase}: This dataset contains 31,000 images primarily focused on human activities, each paired with five descriptive captions. Similar to MS COCO, we adopt bidirectional retrieval metrics to assess cross-modal alignment.
    
    \item \textbf{More General Image Classification Benchmarks}: To further assess generalization capabilities, we evaluate our models on several widely used image classification benchmarks, including \textbf{FER2013}, \textbf{ImageNet-1k}, \textbf{ImageNet-A}, \textbf{ImageNet-O}, \textbf{ImageNet-R}, \textbf{VOC2007}, \textbf{CIFAR-10}, and \textbf{CIFAR-100}. These datasets collectively cover a broad spectrum of visual tasks, such as facial expression recognition, large-scale object classification, out-of-distribution robustness, multi-label classification, and both coarse- and fine-grained category recognition. Empirical results demonstrate that CultureCLIP consistently maintains, and in some cases slightly improves, performance on these benchmarks. This suggests that the incorporation of cultural fine-tuning does not compromise general vision-language alignment or classification capabilities but rather enhances overall robustness and versatility. A comprehensive summary of these results, including Top-1 Accuracy (\textit{Acc1}), Top-5 Accuracy (\textit{Acc5}), and Mean Per-Class Recall (\textit{MPCR}), is provided in Table~\ref{tab:general_benchmarks}.
\end{itemize}

\begin{table*}[h]
\centering
\caption{Performance on general image classification benchmarks (\%). We report Top-1 Accuracy (Acc1), Top-5 Accuracy (Acc5), and Mean Per-Class Recall (MPCR) across various datasets. CultureCLIP maintains or slightly improves general performance despite additional cultural training.}
\label{tab:general_benchmarks}
\resizebox{\textwidth}{!}{
\begin{tabular}{lcccccccc}
\toprule
\textbf{Model} & \textbf{FER2013} & \textbf{ImageNet-1k} & \textbf{ImageNet-A} & \textbf{ImageNet-O} & \textbf{ImageNet-R} & \textbf{VOC2007} & \textbf{CIFAR-10} & \textbf{CIFAR-100} \\
\midrule
\multicolumn{9}{l}{\textit{Top-1 Accuracy (Acc1)}} \\
OpenAI CLIP                      & 41.22 & 63.37 & 31.51 & 47.55 & 69.31 & 76.45 & 89.77 & 64.24 \\
Caption (w/o neg)               & 41.57 & 63.37 & \textbf{31.61} & \textbf{47.80} & 69.28 & \textbf{76.52} & 89.84 & \textbf{64.48} \\
Caption (w/o neg) + Concept (w/o neg) & 41.20 & 63.35 & 31.59 & 47.65 & \textbf{69.33} & 76.42 & \textbf{89.86} & 64.40 \\
Caption (w/ neg)                & \textbf{41.59} & \textbf{63.39} & \textbf{31.61} & 47.65 & 69.26 & \textbf{76.52} & 89.80 & 64.45 \\
CultureCLIP                      & 41.26 & 63.37 & 31.60 & 47.65 & 69.31 & 76.48 & 89.83 & 64.47 \\
\midrule
\multicolumn{9}{l}{\textit{Top-5 Accuracy (Acc5)}} \\
OpenAI CLIP                      & 94.78 & 88.82 & 64.23 & \textbf{78.30} & 88.81 & 95.93 & 99.61 & 88.78 \\
Caption (w/o neg)               & 94.68 & 88.82 & 64.31 & 78.10 & 88.80 & \textbf{95.98} & 99.62 & \textbf{88.91} \\
Caption (w/o neg) + Concept (w/o neg) & \textbf{94.82} & \textbf{88.83} & 64.20 & 78.25 & 88.00 & 95.94 & \textbf{99.63} & 88.84 \\
Caption (w/ neg)                & 94.69 & 88.82 & \textbf{64.33} & 78.10 & 88.79 & 95.96 & \textbf{99.63} & 88.87 \\
CultureCLIP                      & 94.80 & 88.80 & 64.29 & 78.10 & \textbf{88.85} & 95.95 & \textbf{99.63} & 88.85 \\
\midrule
\multicolumn{9}{l}{\textit{Mean Per-Class Recall (MPCR)}} \\
OpenAI CLIP                      & 36.10 & 63.36 & 32.63 & 48.85 & 67.92 & 80.59 & 89.83 & 64.21 \\
Caption (w/o neg)               & 36.40 & 63.36 & 32.75 & \textbf{49.09} & 67.92 & 80.60 & 89.86 & \textbf{64.44} \\
Caption (w/o neg) + Concept (w/o neg) & 36.00 & 63.37 & 32.72 & 49.01 & 67.94 & \textbf{80.64} & \textbf{89.87} & 64.40 \\
Caption (w/ neg)                & \textbf{36.42} & \textbf{63.39} & 32.72 & 49.04 & 67.87 & 80.58 & 89.83 & 64.41 \\
CultureCLIP                      & 36.21 & 63.37 & \textbf{32.77} & 49.00 & \textbf{67.96} & 80.60 & 89.85 & 64.41 \\
\bottomrule
\end{tabular}
}
\end{table*}

\newpage
\section{Error Case Analysis}
\label{app:error cases}
\paragraph{Distinguishing between Gongbi and Xieyi styles}
One illustrative failure case involves differentiating between two classic Chinese painting styles: \textit{gongbi} and \textit{xieyi}. The \textit{gongbi} style is known for its meticulous brushwork, fine lines, and realistic details, often used to depict flowers, birds, and other subjects in a highly controlled and precise manner. In contrast, the \textit{xieyi} style (literally "writing ideas") emphasizes freehand expression, bold strokes, and abstract or suggestive forms rather than realistic details.
In this example, both CLIP and CultureCLIP misclassified an input \textit{gongbi} painting as \textit{xieyi}. However, CultureCLIP assigned a lower confidence to the incorrect label (72\% vs. CLIP's 78\%), indicating a modest calibration improvement. This suggests that while our model still struggles with highly abstract stylistic distinctions, it demonstrates better uncertainty awareness compared to the original CLIP, which is a step toward more nuanced cultural reasoning.

\begin{figure}[h]
  \centering
  \includegraphics[width=0.85\linewidth]{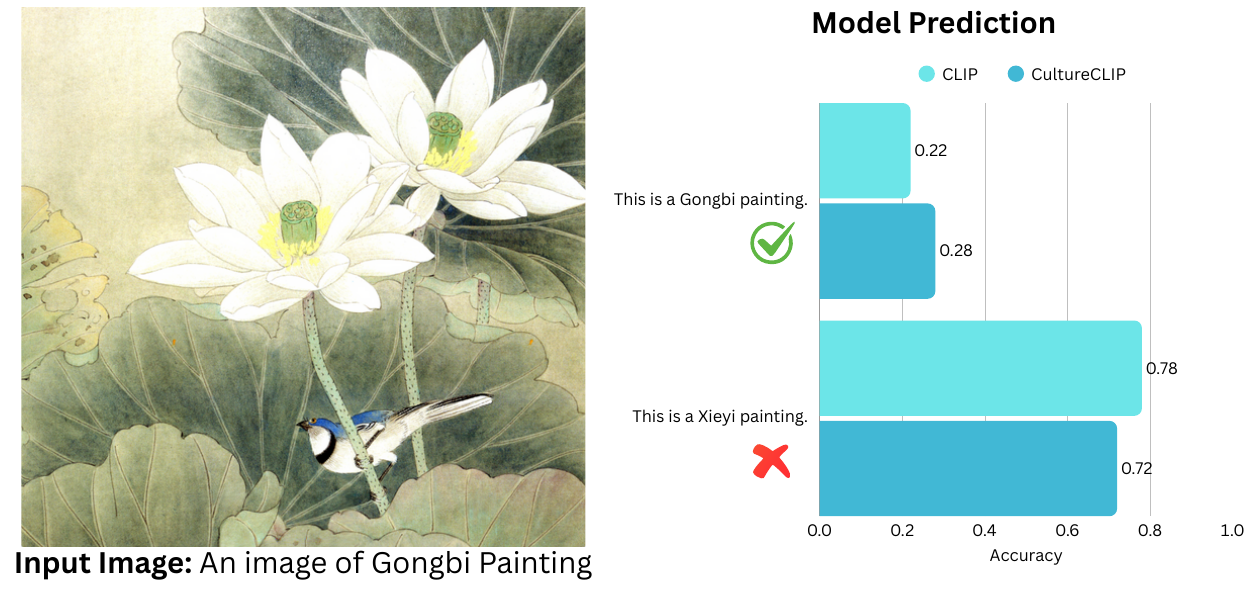}
  \vspace{-0.5em}
  \caption{Example failure case comparing CLIP and CultureCLIP on a \textit{gongbi} painting. While both models misclassified it as \textit{xieyi}, CultureCLIP exhibited lower confidence in its wrong prediction, suggesting better calibration.}
  \label{fig:gongbi_case}
\end{figure}

\section{Prompt Engineering}
\label{app:prompts}
\begin{figure*}[ht]
    \begin{tcolorbox}[colframe=blue!70!black, colback=blue!10!white, title=Bottom-up Filtering Prompt for Qwen2.5-VL, arc=4mm, boxrule=1mm]
    \begin{lstlisting}[style=plain]
Please determine whether the concept '{title}' clearly and unambiguously belongs to one of the eight cultural categories (Cuisine, Clothing, Animals & Plants, Art, Architecture, Daily Life, Symbol, or Festival). If the concept is only loosely related, culturally ambiguous, or does not strongly align with any of the categories, please select 'A' to ensure strict filtering. The concept does not clearly belong to any of the eight categories. The concept is clearly and directly related to one of the eight categories. Answer only with 'A' or 'B'.

Output the result in JSON format as follows:
{
    "concept_type": "A" or "B",
}
    \end{lstlisting}
    \end{tcolorbox}
    \caption{Bottom-up filtering prompt for Qwen2.5-VL.}
    \label{fig:bottom_down_filtering}
\end{figure*}

\begin{figure*}[ht]
    \begin{tcolorbox}[colframe=blue!70!black, colback=blue!10!white, title=Bottom-up Classification Prompt for Qwen2.5-VL, arc=4mm, boxrule=1mm]
    \begin{lstlisting}[style=plain]
Given a cultural concept, definition, definition caption, and definition image, your task is to classify and extract the following information in English:
    - Country: The country or region associated with the concept.
    - Cultural Category: IMPORTANT - You MUST choose exactly ONE category from these eight predefined categories:
        * Cuisine
        * Clothing
        * Animal & Plants
        * Art
        * Architecture
        * Daily Life
        * Symbol
        * Festival
        Do not use any other categories. If the concept doesn't clearly fit into one of these categories, choose the closest match.
    - Context: A short 20-word description in English that provides insight into the cultural and functional use of the concept.
    - Key Visual Features: The visual features that distinguish the concept (e.g., shape, material, color, size) described in English.

Examples:
    Input:
        Concept: Kimono
        Definition: A traditional Japanese garment with long, wide sleeves and an obi sash, worn for formal occasions and festivals.
        Definition Caption: A colorful kimono with crane patterns, displayed at Kyoto's textile museum.
        Definition Image: example_image.jpg
    Output:
        Country: Japan
        Category: Clothing
        Context: A traditional Japanese garment with long, wide sleeves and an obi sash, worn for formal occasions and festivals.
        Key Visual Features: Long, wide sleeves, obi sash, colorful patterns, traditional Japanese style.

Current Task:
    Concept: {concept}
    Definition: {definition}
    Definition Caption: {caption}
    Definition Image: {image_url}
    Generate the Country, Cultural Category (MUST be one of the eight predefined categories), Context, and Key Visual Features in English:
    \end{lstlisting}
    \end{tcolorbox}
    \caption{Bottom-up classification prompt for Qwen2.5-VL.}
    \label{fig:bottom_down_classification}
\end{figure*}

\begin{figure*}[ht]
    \begin{tcolorbox}[colframe=blue!70!black, colback=blue!10!white, title=Top-down Generation Prompt for Qwen2.5-VL, arc=4mm, boxrule=1mm]
        \begin{lstlisting}[style=plain]
Given a country and cultural category, your task is to generate the following:
    - Concept: A cultural concept that fits the provided country and category.
    - Context: A short 20-word description that provides insight into the cultural and functional use of the concept.
    - Key Visual Features: The visual features that distinguish the concept (e.g., shape, material, color, size).
    
Examples:
    Input:
        Country: China
        Cultural Category: Food
    Output:
        Concept: Mantou
        Context: Steamed wheat bun symbolizing prosperity and wisdom, commonly served during festivals and family gatherings.
        Key Visual Features: Pillowy white appearance, round or rectangular shape, smooth surface, typically palm-sized.
    Input:
        Country: Japan
        Cultural Category: Art
    Output:
        Concept: Ukiyo-e
        Context: Traditional woodblock prints depicting scenes from everyday life, nature, and historical events.
        Key Visual Features: Flat color blocks, bold outlines, vibrant pigments, rectangular format, detailed patterns.

Current Task:
    Country: {country}
    Cultural Category: {category}
    Generate the concept, context, and key visual features:
        \end{lstlisting}
    \end{tcolorbox}
    \caption{Top-down generation prompt for Qwen2.5-VL.}
    \label{fig:top_down_concept_generation}
\end{figure*}

\begin{figure*}[ht]
    \begin{tcolorbox}[colframe=blue!70!black, colback=blue!10!white, title=Twin Matching Prompt for Qwen2.5-VL, arc=4mm, boxrule=1mm]
        \begin{lstlisting}[style=plain]
Given a cultural concept from a certain category, its context, and its visual features, your task is to generate the following:
    - New Concept: A visually similar but culturally different concept from the same category.
    - New Context: A short 20-word description in English that provides insight into the cultural and functional use of the generated concept.
    - New Key Visual Features: The visual features that distinguish the concept (e.g., shape, material, color, size) from the original concept.

Examples:
    Input:
        Category: Art
        Concept: Erhu
        Context: A two-stringed Chinese musical instrument played with a bow, often used in traditional Chinese music.
        Key Visual Features: Two strings, a bow, a wooden body, and a horsehair bow.
    Output:
        New Concept: Guzheng
        New Context: A Chinese zither-like instrument with a large, rectangular wooden body and multiple strings, played with plucking.
        New Key Visual Features: A large, rectangular wooden body, multiple strings, and a plucking mechanism.

Current Task:
    Category: {category}
    Concept: {concept}
    Context: {context}
    Visual Features: {visual_features}
    Generate a new concept, context, and visual features:
        \end{lstlisting}
    \end{tcolorbox}
    \caption{Twin matching prompt for Qwen2.5-VL.}
    \label{fig:twin_matching}
\end{figure*}

\begin{figure*}[ht]
    \begin{tcolorbox}[colframe=blue!70!black, colback=blue!10!white, title=Diverse Caption Generation Prompt for Qwen2.5-VL, arc=4mm, boxrule=1mm]
        \begin{lstlisting}[style=plain]
Given a cultural concept, context, and key visual features, your task is to generate 10 different captions that describe the concept in various scenarios while preserving its cultural significance. The captions must reflect different styles or settings, but they should all clearly include the key visual features. Follow these guidelines:
    - Emphasize the key visual differences in the scenes (e.g., shape, size, cooking method, setting).
    - Retain the cultural or functional context (e.g., everyday use, ceremonial purpose, tradition).
    - Ensure the differentiating features (e.g., shape, texture, size, material, use) are clearly reflected in each caption.
    - Each caption should be under 15 words.
    - Each caption should be unique, showing different perspectives or settings, but should always include the key visual features.

Examples:
    Input:
        Concept: Xiaolongbao (Soup Dumplings)
        Context: Steamed soup dumplings with delicate, thin wrappers, filled with savory broth and pork, typically steamed in bamboo baskets.
        Key Visual Features: Delicate, thin wrappers and steamed in bamboo baskets.
    Output:
        1. A chef carefully places Xiaolongbao in a bamboo steamer, showcasing their thin, translucent wrappers and savory broth inside.
        2. A steaming bamboo basket of Xiaolongbao, delicate wrappers holding savory broth, served in an elegant Shanghai restaurant.
        3. Steamed xiaolongbao resting in bamboo baskets, ready to be served during a family meal.
        4. Crispy fried xiaolongbao, golden-brown and served with dipping sauce, sitting in bamboo baskets.
        5. Miniature xiaolongbao filled with crab roe, elegantly presented in bamboo baskets at a Cantonese restaurant.
        6. Steaming xiaolongbao with delicate skin, served in bamboo baskets during a traditional Chinese New Year meal.
        7. Bamboo-steamed xiaolongbao, filled with savory broth, served alongside hot tea in a Beijing teahouse.
        8. Translucent, plump xiaolongbao, freshly steamed in bamboo baskets for a cozy brunch setting.
        9. Steamed xiaolongbao with pork filling, served in bamboo baskets with chili oil at a street food stall.
        10. Elegant xiaolongbao, arranged in bamboo baskets, presented at a lavish festive feast.

Current Task:
    Concept: {concept}
    Context: {context}
    Key Visual Features: {visual_features}
    Generate 10 different captions, each reflecting a different style or scene, but all incorporating the key visual features:
        \end{lstlisting}
    \end{tcolorbox}
    \caption{Diverse caption generation prompt for Qwen2.5-VL.}
    \label{fig:caption_diversification_prompt}
\end{figure*}

\begin{figure*}[ht]
    \begin{tcolorbox}[colframe=blue!70!black, colback=blue!10!white, title=Prompt for Quality Evaluation on Authenticity, arc=4mm, boxrule=1mm]
        \begin{lstlisting}[style=plain]
Please analyze this image and rate its authenticity on a scale of 1 to 5. You can refer to the context to help you make the decision. Focus on whether the concept shown is realistic and follows common sense. Consider:
    - Are all elements anatomically and physically correct?
    - Does everything look natural and possible in the real world?
    - Are there any unrealistic or deformed features?

Examples:
    Input: 
        Concept: Erhu
        Context: The Erhu is a traditional Chinese musical instrument played with a bow. It consists of a wooden body and two strings, and is known for its expressive and resonant sound. Often used in Chinese classical and folk music, it is typically performed in both solo and ensemble settings.
        Image: An image showing a person playing Erhu with three hands in mid air
    Output: 1
    Input: 
        Concept: Erhu
        Context: The Erhu is a traditional Chinese musical instrument played with a bow. It consists of a wooden body and two strings, and is known for its expressive and resonant sound. Often used in Chinese classical and folk music, it is typically performed in both solo and ensemble settings.
        Image: An image showing a person playing Erhu with two hands sitting on the chair;
    Output: 5

Now, give you the image, concept and its corresponding context:
    Concept: {concept}
    Context: {context}
    Output only the score:
        \end{lstlisting}
    \end{tcolorbox}
    \caption{Prompt for quality evaluation on authenticity.}
    \label{fig:authenticity_quality}
\end{figure*}

\vspace{0.3cm}

\begin{figure*}[ht]
    \begin{tcolorbox}[colframe=blue!70!black, colback=blue!10!white, title=Prompt for Quality Evaluation on Consistency, arc=4mm, boxrule=1mm]
        \begin{lstlisting}[style=plain]
Please analyze this image and rate its consistency with the concept on a scale of 1 to 5. You can refer to the context to help you make the decision. Focus on whether the image accurately depicts the specified concept without showing wrong concepts. Consider:
    - Does the image show exactly the concept mentioned?
    - Are there any mismatched or wrong elements?
    - Is the concept clearly and accurately represented?

Examples:
    Input: 
        Concept: Tang Sancai
        Context: A type of Chinese glazed pottery from the Tang Dynasty, known for its colored glazes (green, yellow, and brown) often used for tomb figurines and decorative pieces.
        Image: An image showing a blue-and-white porcelain bowl
    Output: 1
    Input: 
        Concept: Tang Sancai
        Context: A type of Chinese glazed pottery from the Tang Dynasty, known for its colored glazes (green, yellow, and brown) often used for tomb figurines and decorative pieces.
        Image: An image showing exactly a Tang Sancai horse
    Output: 5

Now, give you the image, concept and its corresponding context:
    Concept: {concept}
    Context: {context}
    Output only the score:
        \end{lstlisting}
    \end{tcolorbox}
    \caption{Prompt for quality evaluation on consistency.}
    \label{fig:consistency_quality}
\end{figure*}

\vspace{0.3cm}

\begin{figure*}[ht]
    \begin{tcolorbox}[colframe=blue!70!black, colback=blue!10!white, title=Prompt for Quality Evaluation on Cultural Fidelity, arc=4mm, boxrule=1mm]
        \begin{lstlisting}[style=plain]
Please analyze this image and rate its cultural fidelity on a scale of 1 to 5. Focus on whether the cultural elements are accurate and appropriate for the specific context. Consider:
    - Are all cultural elements accurate for this context?
    - Are there any mixed or incorrect cultural elements?
    - Does everything align with the cultural background specified?

Examples:
    Input: 
        Concept: Mexican Day of the Dead
        Context: A Mexican holiday celebrating deceased loved ones, featuring marigold flowers, sugar skulls, and candle altars. 
        Image: An image showing marigold flowers, skull paintings, and candle altars.
    Output: 5
    Input: 
        Concept: Mexican Day of the Dead
        Context: A Mexican holiday celebrating deceased loved ones, featuring marigold flowers, sugar skulls, and candle altars. 
        Image: An image showing showing marigold flowers, skull paintings, and Chinese joss paper money.
    Output: 1

Now, give you the image, concept and its corresponding context:
    Concept: {concept}
    Context: {context}
    Image:{image}
    Output only the score:
        \end{lstlisting}
    \end{tcolorbox}
    \caption{Prompt for quality evaluation on cultural fidelity.}
    \label{fig:cultural_fidelity_quality}
\end{figure*}

\end{document}